\documentclass[12pt]{article}

\usepackage{amsmath,amsthm, amsfonts, amssymb, amsxtra, amsopn}
\usepackage{pgfplots}
\usepgfplotslibrary{colormaps}
\pgfplotsset{compat=1.18}
\usepackage{pgfplotstable}
\usetikzlibrary{pgfplots.statistics}
\usepackage{graphicx,grffile}
\usepackage{multirow}
\usepackage{booktabs}
\usepackage{listings}
\usepackage{cmap}
\usepackage{colortbl}
\usepackage{adjustbox}
\usepackage{epsfig}
\usepackage[framemethod=tikz]{mdframed}

\newmdenv[tikzsetting={draw=black,fill=black,fill opacity=0.175}]{mymdframed}

\usepackage[tableposition=top,font=small,skip=5pt]{caption}
\usepackage{subcaption}
\usepackage{makecell} 

\pgfkeys{
    /pgf/number format/fixed zerofill=true }
    
\pgfplotstableset{
    /color cells/min/.initial=0,
    /color cells/max/.initial=1000,
    /color cells/textcolor/.initial=,
    color cells/.code={%
        \pgfqkeys{/color cells}{#1}%
        \pgfkeysalso{%
            postproc cell content/.code={%
                \begingroup
                \pgfkeysgetvalue{/pgfplots/table/@preprocessed cell content}\value
\ifx\value\empty
\endgroup
\else
                \pgfmathfloatparsenumber{\value}%
                \pgfmathfloattofixed{\pgfmathresult}%
                \let\value=\pgfmathresult
                \pgfplotscolormapaccess
                    [\pgfkeysvalueof{/color cells/min}:\pgfkeysvalueof{/color cells/max}]%
                    {\value}%
                    {\pgfkeysvalueof{/pgfplots/colormap name}}%
                \pgfkeysgetvalue{/pgfplots/table/@cell content}\typesetvalue
                \pgfkeysgetvalue{/color cells/textcolor}\textcolorvalue
                \toks0=\expandafter{\typesetvalue}%
                \xdef\temp{%
                    \noexpand\pgfkeysalso{%
                        @cell content={%
                            \noexpand\cellcolor[rgb]{\pgfmathresult}%
                            \noexpand\definecolor{mapped color}{rgb}{\pgfmathresult}%
                            \ifx\textcolorvalue\empty
                            \else
                                \noexpand\color{\textcolorvalue}%
                            \fi
                            \the\toks0 %
                        }%
                    }%
                }%
                \endgroup
                \temp
\fi
            }%
        }%
    }
}

\let\subsubsubsection=\paragraph

\PassOptionsToPackage{hyphens}{url}

\usepackage{hyperref}
\hypersetup{colorlinks=true,linkcolor=black,citecolor=black,urlcolor=blue,filecolor=black}
\hypersetup{pdfpagemode=UseNone,pdfstartview=}

\usepackage{enumitem}
\setlist[itemize]{noitemsep, topsep=0pt}

\advance\oddsidemargin by -0.35in
\advance\textwidth by 0.7in

\advance\topmargin by -0.4in
\advance\textheight by 0.8in

\long\def\symbolfootnotetext[#1]#2{\begingroup%
\def\thefootnote{\fnsymbol{footnote}}\footnotetext[#1]{#2}\endgroup}

\newcommand\dunderline[3][-1pt]{{%
  \sbox0{#3}%
  \ooalign{\copy0\cr\rule[\dimexpr#1-#2\relax]{\wd0}{#2}}}}
\def\uuu{\kern-1pt\dunderline{0.75pt}{\phantom{M}}}

\title{Distinguishing Chatbot from Human}

\author{Gauri Anil Godghase\footnotemark[1]\ \ \ 
Rishit Agrawal\footnotemark[1]\ \ \
Tanush Obili\footnotemark[1]\ \ \
Mark Stamp\footnotemark[1]\,\,\footnotemark[2]}

\begin{document}

\symbolfootnotetext[1]{Department of Computer Science, San Jose State University}
\symbolfootnotetext[2]{mark.stamp$@$sjsu.edu}

\maketitle

\abstract
There have been many recent advances in the fields of generative Artificial Intelligence (AI) 
and Large Language Models (LLM), 
with the Generative Pre-trained Transformer (GPT) model being a leading ``chatbot.'' 
LLM-based chatbots have become so powerful 
that it may seem difficult to differentiate between human-written and machine-generated text. 
To analyze this problem, 
we have developed a new dataset consisting of more than~750,000 human-written paragraphs,
with a corresponding chatbot-generated paragraph for each. Based on this dataset,
we apply Machine Learning (ML) techniques to determine the origin of text (human or chatbot). 
Specifically, we consider two methodologies for tackling this issue: feature analysis 
and embeddings. Our feature analysis approach involves extracting a collection of features from the 
text for classification. We also explore the use of contextual embeddings and transformer-based architectures 
to train classification models. Our proposed solutions offer high classification accuracy and serve as useful tools for 
textual analysis, resulting in a better understanding of chatbot-generated text in this era of advanced AI technology.

\section{Introduction}\label{introduction}

Recent advances in Large Language Models (LLM) have forever changed the field of 
Natural Language Processing (NLP). A front-runner in the LLM industry has been the 
Generative Pre-trained Transformer (GPT)~\cite{radford2018improving} series of models,\footnote{At
the time that this research was initiated, the state-of-the-art version of the GPT model was
GPT-3.5, and hence that is the version that we use in all experiments discussed in this paper.} 
colloquially known as ChatGPT.
These GPT models are known for their large scale, parameter size, advanced language processing 
abilities and creative text generation. The prominence of GPT in the current field makes this model an 
important topic of research. The GPT model, and other LLMs are gradually becoming 
more human-like. The line between human written text and LLM-generated text is likely to 
continue to blur. This advent of LLMs, although beneficial in several fields, also presents challenges 
associated with discerning the origin of written text. 

In this research, we explore the use of Machine Learning (ML) 
and Deep Learning (DL) techniques for the classification of text into two categories: 
human-written and GPT-generated. This classification is important for several domains, including content moderation, 
cybersecurity, education, and so on.

We first collect a large dataset of more than~750,000 human-written paragraphs, each of which includes 
a brief summary. Then, for each of these human-written paragraphs, we ask ChatGPT to generate a corresponding paragraph 
on the same topic and of approximately the same length, which yields a new, high-quality dataset containing more than~1,500,000
full paragraphs. We then determine how accurately we can classify these text samples as human 
or chatbot. For the classification task, we consider the following two approaches. 
\begin{itemize}
\item  Feature analysis --- Our feature analysis approach involves extracting a wide range of features from 
the data samples and using elementary statistical properties of these features to classify text as human or chatbot generated. 
This involves analyzing the lexical diversity, linguistics, syntactical structures, and other characteristics of the data.
\item Embeddings --- Our embeddings approach involves feeding data to learning models, based on word and sentence embeddings. 
These embeddings are vectors that are designed to capture some relationships present in the data, enabling 
models to learn from the underlying semantics of the text.
\end{itemize}

At the core of our research are various ML and DL techniques that we use to classify text. 
Our main goal is to determine how accurately such learning techniques are able to distinguish human-generated text from 
GPT-generated text. Another important aspect of this research is to determine which features are most useful in this classification. 
Understanding the relative importance of various various features will allow us to better understand the strengths and limitations of 
ChatGPT-generated content and to identify areas for improvement.

The remainder of this paper is organized as follows. 
Section~\ref{chap:previous} explores the existing literature surrounding this topic, 
while in Section~\ref{background} we focus on the background required
to understand our work. Section~\ref{dataset} covers the process we follow for dataset generation, 
and Section~\ref{data exploration} presents some elementary statistical analysis of our new dataset. 
Section~\ref{implementation} delves into implementation details of our learning-based approaches for 
distinguishing between human and chatbot text, including the features extracted, 
embedding techniques, the learning models that we train, and so on. 
In Section~\ref{results} we discuss the results that we have obtained. The
paper concludes with Section~\ref{conclusion}, where we also discuss 
some potential avenues for future work.

\section{Relevant Related Work}\label{chap:previous}

In this section, we first review the chatbot that we use to generate our data, namely, ChatGPT.
Then we delve into the existing literature and research studies that have explored the differentiation between 
human and machine-generated text. 
This review of previous work serves to place our research in context, as of the time this paper was written. 
However, it is worth noting that this is a rapidly evolving field, and additional new research and results
are certain to appear.

\subsection{ChatGPT}

The GPT~3.5 model used by ChatGPT, developed by OpenAI, is a deep learning model based 
on a transformer architecture. It is a pre-trained model which has been trained on a vast corpus of information 
from the Internet, and several other publicly available sources, including books, websites, 
etc.~\cite{radford2018improving,RAY2023121}. 
The exact size amount of training data used for this model is not publicly known. However, a previous
version of the model had~175 billion 
parameters and was trained with~499 billion crawled text tokens. The model is able to recognize patterns within text 
and generate information that closely resembles text written by humans. During the pre-training, the model learns 
to predict the next word in a sentence~\cite{RAY2023121}. This is done using an attention-based transformer 
architecture~\cite{attention}, which enables the model to learn contextual information and patterns within the text. 
The model first interprets the context of the user's query and autoregressively generates the next word 
(or token) that fits the context. 

\subsection{Human Classification}

Before diving into the classification of text using machine learning, we might consider the question: How good are humans at 
distinguishing between machine and human content? The research in~\cite{10.1007/978-3-030-78635-9_67} 
attempted to answer this question. This study included nine literature professionals and each was given the 
initial lines of~18 poems and short stories by classic authors. They were asked to produce two continuations for each text, 
one with AI tool based on the GPT~2 model. While evaluating, the participants were asked to classify all continuations 
not written by themselves as AI or human. The results of this study indicated that the professional struggled with 
this classification problem, with high rates on false positives and negatives. These results indicate that the problem of 
identification of human versus machine generated content may be moving out of human hands. 
The research in~\cite{ippolito2020automatic} presents evidence suggesting longer excerpts of text can fool humans 
over~30\%\ of the time. Thus, it would appear that we may need machines to accurately distinguish between 
human-generated and chatbot-generated text.

\subsection{Datasets}

To date, research on the classification of text as human or GPT-generated has been scant. There is also a paucity of 
publicly available datasets for experimentation. Most researchers have thus had to create their own datasets
(and this paper is no exception). In the study~\cite{10.1007/978-981-99-7947-9_12}, the authors created a dataset based 
on Wikipedia articles. They defined~10 categories and selected~10 topics within each category to generate content 
using AI. 
Similarly, the authors of~\cite{hayawi2023imitation} developed a dataset based on scientific papers. 
They passed the title of the paper as a prompt and asked the model to generate an abstract based 
on the title. In contrast to using a single dataset for classification, the authors of~\cite{hayawi2023imitation} 
created several categories of data, such as essays, poems, stories and code. All of these were used for 
classification. These papers indicate that a wide array of creative techniques have been used to generate data.

\subsection{Classification Techniques}

The approaches for classification using Machine Learning have also varied widely in published studies. 
The research of~\cite{10.1007/978-981-99-7947-9_12} and~\cite{ma2023ai} both used feature extraction based approach 
for classification. On the other hand, the authors of~\cite{hayawi2023imitation} transformed the textual data to numerical 
using TFIDF\footnote{We discuss TFIDF in Section~\ref{sect:TFIDF}.}
and similar techniques. The authors of~\cite{ippolito2020automatic} used a ``bag of words'' 
implementation to generate embeddings. Thus, two primary methods have been adopted for classification problem:
feature based and embedding based.

\subsection{Feature Based Classification}

The authors of~\cite{10.1007/978-981-99-7947-9_12} extracted~37 features grouped in eight categories.
(perplexity, semantic, list lookup, document, error based, readability, AI feedback, and text vector features). 
On the other hand, the authors of~\cite{ma2023ai} used features in just three categories (syntax, semantics, and pragmatics). 
The syntax features are token level (e.g., length of words, part of speech, function word frequency, and stopword ratio), 
semantic features consist of cosine similarities between sentences (including the title), and pragmatic features 
deal with things like self-contradictions and redundancies.

An interesting approach was adopted in~\cite{10.1007/978-981-99-7947-9_12}, which included asking the GPT model itself 
if specific text was generated by it. Another feature category in this paper was perplexity, which is a measure of how 
surprised the language model is when it encounters a new sequence of words. The remaining features 
in~\cite{10.1007/978-981-99-7947-9_12} were similar to~\cite{ma2023ai}.

\subsection{Model Selection}

Several different models have been used for text classification in various studies,
with Logistic Regression being often used to establish baseline metric scores. 
The work in~\cite{hayawi2023imitation,ippolito2020automatic,ma2023ai} 
all use Logistic Regression for classification; in~\cite{hayawi2023imitation,ippolito2020automatic} 
it is used for classification of text into human or AI-generated content, whereas 
in~\cite{ma2023ai} it is used to interpret the various features. 
In~\cite{hayawi2023imitation,ippolito2020automatic}, 
Logistic Regression yielded an average accuracy of~0.79 to~0.93 in various cases. 

XGBoost, Random Forest, and Multilayer Perceptron models were used for classification 
in~\cite{hayawi2023imitation}, where an accuracy of~0.98 was attained for basic AI-generated texts 
and~0.969 for more advanced cases. Bidirectional Encoder Representations from Transformers (BERT)
was used in~\cite{ippolito2020automatic}. The authors of~\cite{hayawi2023imitation} experimented 
with the Long Short Term Memory (LSTM) models.

\section{Background}\label{background}

In this research, we attempt to classify text as either human or GPT~3.5 generated using 
Machine Learning and Deep Learning techniques. As mentioned above, we consider two approaches,
namely, feature analysis and embeddings. In this section, we explain in some detail background 
topics, including the various embedding techniques and the learning models used.

\subsection{Models}\label{background:models}

In this section, we provide an overview of all of the ML and DL models that were used in the project. 
Among classic ML techniques, we consider Logistic Regression, Random Forest, and XGBoost.,
while from DL models, we experiment with Multilayer Perceptron, a Deep Neural Network, 
and Long Short Term Memory networks. 

\subsubsection{Logistic Regression}

Logistic Regression (LR) models the relationship between independent and dependent variables. 
LR is useful when we need to predict the possibility of the occurrence of an event, 
and it is most often used for binary classification problems~\cite{boateng2019review}.

LR predicts the likelihood of an observation belonging to a particular class by using the logistic or the sigmoid function 
to map the probability of outcomes to the range~0 to~1. It assumes 
that a linear relationship exists between the predictor variable and the log odds of the feature 
variables~\cite{analyticsvidhya_logistic_regression}. 

\subsubsection{Random Forest}

Random Forest (RF) is one of the most popular supervised ML algorithms. It is an ensemble technique 
that is trained using a ``bagging'' approach. In bagging, multiple weak learners (decision trees in the case of RF) 
are trained in parallel, each based on a subset of the features and data. 
Classification of individual data points into classes is based on the ensemble
of these weak learners~\cite{stampML}. 
One advantages of RF is that it can effectively handle missing values. 

\subsubsection{Support Vector Machine}\label{sect:SVM}

For a binary classification problem, Support Vector Machines (SVM) attempt to
separate the classes using a hyperplane, while maximizing the ``margin,'' i.e., the
minimum distance between the hyperplane and the training data.
The so-called kernel trick enables us to embed a nonlinear transformation
into the SVM training process---which can serve to increase the separation between
classes---without any significant loss of efficiency. In this paper, we only consider 
linear SVMs, in which case each feature has an associated weight which specifies
the importance that the model places on that specific feature~\cite{stampML}.

\subsubsection{XGBoost}

XGBoost is short for eXtreme Gradient Boosting. Like RF, 
it is also an ensemble learning model that uses various weak learners (trees) to give predictions. 
However, boosting techniques rely on an involved process for combining weak learners,
as opposed to the simple voting strategy of an RF.
XGBoost employs a block structure for parallel learning, enabling efficient 
distributed computing~\cite{Chen_2016}.

XgBoost is known for its scalability. It is capable of handling datasets that scale beyond billions of 
examples. However, potential disadvantage of XGBoost include that 
it is prone to overfitting and sensitive to outliers.

\subsubsection{Multilayer Perceptron}

Multilayer Perceptron (MLP) is a type of feedforward Artificial Neural Network (ANN). 
An MLP can effectively deal with nonlinear relationships within the data. MLPs are known for their 
applications in a wide range of domains, including natural language processing (NLP). Within the NLP domain, 
MLPs have been successfully used for various tasks, including machine 
translation and speech recognition. 

The architecture of an MLP consists of multiple layers of neurons with each layer being 
fully connected to the next~\cite{stampML}. There are three types of layers present---one input layer, 
one output layer, and a small number of hidden layers, where ``small'' is typically one or two. 
The number of neurons in each layer and the number of hidden layers are parameters that need 
to be determined experimentally. Each neuron has a nonlinear activation function 
associated with it, such as a sigmoid, rectified linear unit (ReLU), 
or absolute value activation function~\cite{yu2023multilayer}.

\subsubsection{Deep Neural Network}

In our usage,
Deep Neural Networks (DNN) are a more general form of ANN,
with an architecture similar to MLP,  but with a larger number of hidden layers. 
In addition, a DNN may contain different types of layers, such as convolutional layers,
pooling layers, or recurrent layers~\cite{9075398}. 
The presence of these layers helps in distinguishing between different types of DNNs,
such as Convolutional Neural Networks (CNN) and Recurrent Neural Networks (RNN).
For the experiments considered in this paper, the DNN architecture is just a deeper version 
(i.e., more hidden layers) of an MLP.

Training of deep neural networks involves adjusting the weights between layers to minimize 
the error between input and output. This is most efficiently done by a process known as 
backpropogation~\cite{stampML}. This process is iteratively repeated 
over the training data, with the trained version of the model then used to make predictions on 
previously unseen data. DNNs have proven useful for handling complex data and they
generally give good performance on such datasets. However, they require a large amount 
of data for training and the models themselves are notoriously difficult 
to interpret~\cite{stampML}.

\subsubsection{Long Short Term Memory}

Long Short Term Memory (LSTM) networks are a special class of 
Recurrent Neural Networks (RNN)~\cite{otter2021survey}. 
RNNs are ANNs that possess some ``memory,'' in the sense that they can use data from previous
time steps to make decisions. RNNs have many applications in NLP that require context, such
as predicting the next word in a sentence or machine translation. However, generic RNNs
tend to suffer from gradient pathologies (e.g., vanishing or exploding gradient) when trained 
via backpropagation, making it difficult to effectively use information that is farther
back in time. LSTMs mitigate these gradient issues by use of a complex gating structure,
which enables information to flow more easily through multiple time steps. For additional
details on LSTMs, see~\cite{stampML,vennerod2021long}.

\subsection{Word Embeddings}\label{sect:embed}

Word embeddings are a powerful concept in NLP that allow computers to understand and 
manipulate words based on their meanings. Word embeddings are numerical representations of 
words represented by vectors. They are designed to enable models to process the nuances of 
language, similar to the way that humans do. They consist of multi-dimensional arrays where 
each word is mapped to a vector in a predefined vector space. The goal of embeddings is to 
place similar words closer together (in some well-defined sense) within the vector space.

\subsubsection{TFIDF}\label{sect:TFIDF}

Term Frequency Inverse Document Frequency (TFIDF) is simple method to generate numerical representations 
of words, that was originally developed to automatically extract indexing terms from documents. 
Term Frequency (TF) measures how frequently a word occurs in a document.
Specifically,
$$
  \text{TF}(t, d) = \frac{n_{t,d}}{\displaystyle\sum_{w \in d} n_{w,d}} 
$$
where~$n_{t,d}$ is the number of times term~$t$ appears in document~$d$ and
$$
  \sum_{w \in d} n_{w,d} 
$$
is the total number of words in document $d$. A higher TF score simply means that a word occurs more frequently 
in a text. On the other hand, Inverse Document Frequency (IDF) measures how important a term is. 
It penalizes words that occur too frequently across all documents~\cite{stampML}. 
By taking the logarithm of the division, IDF reduces the effect of terms that appear very frequently in the dataset. 
This is done because the terms that occur frequently across all documents are less likely to be informative. 
Specifically, the IDF of a term is calculated by
$$
  \text{IDF}(t, d) = \log\left(\frac{N}{n_t}\right)
$$
where~$N$ is the total number of documents in the collection~$d$ and~$n_{t}$ 
is the number of documents containing the term~$t$.

By multiplying the TF and IDF scores together we obtain the TFIDF score. A higher TFIDF score for a word 
indicates that the term is frequent in a particular document but not so frequent across all the other documents. 
This term is likely to be a distinguishing characteristic of that particular document. .

\subsubsection{Word2Vec}

Word2Vec is a series of related models that are used to produce word embeddings. 
Word2Vec models are designed to capture the syntactic and semantic relationships between 
words~\cite{word2vec}. This is done by placing words in a continuous vector 
space where words with similar meanings are located close (in the sense of cosine similarity). 
to one another.

Word2Vec can be implemented using two different architectures: Continuous Bag-Of-Words (CBOW) 
and Skip-Gram. CBOW predicts a target word based on its context, and it is faster and tends to produce 
better results for more frequent words. Skip-Gram, on the other hand, does essentially the opposite, 
using a target word to predict the surrounding context. It performs well with small datasets and is 
effective at representing rare words~\cite{word2vec}. For the research
reported in this paper, we use the CBOW architecture.

\subsubsection{GloVe}

Global Vectors for Word Representation (GloVe) is an unsupervised learning algorithm designed 
to generate word embeddings. It uses information about the co-occurrence of words within a corpus to generate its
embeddings~\cite{glove}. These embeddings are vector 
representations of the corpus and, and they are dense in the vector space. They capture the semantic meanings 
and relationships between words, which allows the model to understand the nuances of the 
language. 

The core concept of GloVe is to analyze the probabilities 
of word co-occurrences across a text corpus to learn word vectors that reflect collective usage patterns.
This is accomplished by constructing a word-context co-occurrence matrix. This matrix represents the 
frequency with which words in the corpus occur near each other (within a specified context window). 
This matrix is then factorized to lower its dimensionality, using various matrix factorization 
techniques~\cite{glove}. The values of these vectors are iteratively changed and 
optimized to minimize the difference between the co-occurrence probabilities in the original matrix and 
the dot product of the resulting word vectors---this is known as the reconstruction loss. Through this process, 
GloVe captures global word usage patterns, where both semantic and syntactic information is
contained within the vectors.

\subsubsection{BERT}

Bidirectional Encoder Representations from Transformers (BERT) is an extremely popular 
word embedding method. BERT is said to have revolutionized NLP by allowing machines to 
understand and process text with high accuracy~\cite{bert}.
Unlike GloVe and Word2Vec embeddings, BERT embeddings capture more information about 
context.

BERT generates embeddings by processing text in both forward and backwards 
directions. This means that it takes into 
account the words on both sides of the word currently being processed. Before BERT, 
most models considered text only the forward direction. 

BERT has a maximum input limit of 512 tokens. Input words are broken down further into smaller units 
called WordPieces. The model is pre-trained using two strategies: 
Masked Language Modeling (MLM) and Next Sentence Prediction (NSP). 
MLM randomly masks some of the tokens from the input and predicts them based on their context. 
NSP predicts whether two segments of text follow each other in the original document. 
This pre-training allows BERT to achieve a highly develop model of language structure and context.

At the core of the BERT architecture, lie transformers,
which allow for simultaneous processing using a mechanism known 
as ``attention'' to determine which parts of the data are most relevant.
This attention mechanism is based on an encoder-decoder model
that allows BERT to capture the nuances of language and complex sentence 
structures~\cite{attention}. 

\section{Dataset}\label{dataset}

This section outlines the process we used to generate our dataset. We also discuss the numerous features that 
we extract from the data for our feature analysis approach. These features lay a foundation for differentiating between 
the two classes of text.

\subsection{Raw Data Generation}

Our dataset consists of a combination of human-generated and GPT-generated text. For the human-generated text, 
the publicly available WikiHow dataset introduced in~\cite{koupaee2018wikihow} was used. The original dataset 
consists of four columns: \texttt{title}, \texttt{overview}, \texttt{headline}, and \texttt{text}. The \texttt{title} 
consists of the title of the WikiHow article, 
the \texttt{overview} is an introduction, the \texttt{headline} is a bold headline that occurs before the paragraph, 
and, finally, the \texttt{text} paragraph is the actual text of article.

Each WikiHow article in our dataset was used to generate a corresponding GPT paragraph of approximately the same length. 
This was done to maintain the similarity of topics and length of paragraphs between the two sets of data. 
The prompt in Figure~\ref{fig:prompt} was passed to the GPT~3 API, replacing the placeholders with the corresponding information 
from each WikiHow paragraph. Some examples human-generated text and the corresponding GPT-generated text 
can be found in Appendix~\ref{appendix_a}.

\begin{figure}[!htb]
\centering
\begin{mymdframed}[leftmargin=1cm,rightmargin=1cm,roundcorner=5pt]
I am writing an article titled \{\texttt{title}\} for a WikiHow page. 
Write a paragraph of length \{\texttt{length}\} whose title is \{\texttt{headline}\} 
for the \{\texttt{sectionLabel}\} section of this article.
\end{mymdframed}
\caption{ChatGPT prompt}\label{fig:prompt}
\end{figure}

The resulting response from the GPT~3.5 model was extracted and all of the responses for all of the WikiHow
articles in our dataset comprise the GPT generated data. The structure of the prompt was intended to ensure that the content of 
the GPT-generated data mirrored the content of the human data in information and length. Our goal is to ensure that the 
models we consider differentiate the text based on the fundamental characteristics, rather than simply based on topics 
or other extraneous aspects.  Once the chatbot data was generated, the two sets were merged. 
The resulting dataset is balanced, with
the same number of human and chatbot-generated samples.

Basic statistics for our dataset appear in Table~\ref{tab:dataset}. 
It is immediately apparent that ChatGPT tends to be ``wordy,'' in comparison to human writers.
Note also that ChatGPT sometimes produced responses consisting of more than one paragraph, 
and hence we obtain more ChatGPT-generated paragraphs than the number of human-generated paragraphs. 

\begin{table}[!htb]
\centering
\caption{Dataset details}\label{tab:dataset}
\adjustbox{scale=0.85}{
\begin{tabular}{c|cccc}
\toprule
\multirow{2}{*}{\textbf{Class}} & \multirow{2}{*}{\textbf{Paragraphs}} 
	& \multirow{2}{*}{\textbf{Words}} & \multirow{2}{*}{\textbf{Characters}} & \textbf{Average words} \\ 
 & & & & \textbf{per paragraph} \\ \midrule
Human & 784,636 & 54,005,604 & 307,005,548 & 68.83  \\ 
ChatGPT & 920,259 & 75,474,378 & 474,396,685 & 82.01 \\ \bottomrule
\end{tabular}
}
\end{table}

\subsection{Features}\label{sect:FA}

This section covers the various types of features that were generated for every paragraph of data. We briefly discuss
the four broad categories of features that we consider, and even more briefly introduce each of the individual within
these categories.

\subsubsection{Linguistic Features}

Our linguistic features are designed to capture information about an author's unique voice and approach to language. 
It includes the choice of words, their arrangement, and the use of various parts of speech. These features convey meaning, 
tone, and personality. This category is used to understand how language expresses ideas and emotions. These features 
combined contribute to the readability and nuances of the text. Linguistic style is one of the fundamental aspects of text analysis,
and it offers insights into the texture and flavor of the language. 

Note that when creating these linguistic features, all values were normalized and ratios were calculated, instead of using raw 
frequencies. This was done to ensure minimal influence of lengths of the text and to minimize further data preprocessing. 
Next, we introduce each the eight linguistic features that we consider.

\subsubsubsection{Verb ratio}\!\!\!---
The verb ratio refers to the ratio of the frequency of verbs in a text to the total number of words. 
Verbs refer to the action words in the text. They are essential in the construction of sentences, 
conveying actions and states. A higher verb ratio indicates text that is more dynamic. 

\subsubsubsection{Noun ratio}\!\!\!---
The noun ratio is the ratio of the frequency of nouns in the text to the total number of words. 
Nouns are the fundamental building blocks of sentences consisting of the names of people, places, things, or ideas. 
A text with a high noun ratio indicates richness of the text, in terms of subjects and concepts and abstract entities,
and suggests that the text is dense with information and ideas.

\subsubsubsection{Adjective ratio}\!\!\!---
The adjective ratio is the ratio of the frequency of adjectives within a text to the total number of words. 
Adjectives describe or modify nouns. They provide more refined information about the qualities, quantities, 
or states of being of the nouns in the sentence. A higher ratio of adjectives indicates a more descriptive or expressive text,
and create a more detailed picture of the corresponding nouns.

\subsubsubsection{Pronoun ratio}\!\!\!---
The pronoun ratio is the ratio of the frequency of pronouns to the number of words. 
Pronouns are used to replace nouns. A higher pronoun ratio indicates frequent references to 
previously occurring nouns, which tends to make the text more personalized and generally easier to read and follow.

\subsubsubsection{Adverb ratio}\!\!\!---
The adverb ratio is the ratio of the frequency of adverbs within a text to the total number of words. 
Adverbs describe or modify verbs, adjectives, or other adverbs. They provide more information on how, 
when, where, and to what extent actions are performed. Text with a high adverb ratio provides more detailed 
descriptions of how actions are performed, and they add to the depth of the text, in the sense of making it more descriptive.

\subsubsubsection{Preposition ratio}\!\!\!---
The preposition ratio is the ratio of the frequency of prepositions within a text to the total number of words. 
Prepositions are words that link nouns, pronouns, or phrases to other words within a sentence. 
They indicate temporal, spatial or other relationships of objects. A text with a high preposition ratio 
contains more complex descriptions of places, times, and other relationships.

\subsubsubsection{Conjunction ratio}\!\!\!---
The conjunction ratio is the ratio of the frequency of conjunctions within a text to the total number of words. 
Conjunctions join together different words, phrases, clauses, or sentences. They allow a seamless flow of 
words and ensure the coherence of the text. A higher conjunction ratio indicates complex and interwoven 
ideas, and can enhance the reader's ability to follow these ideas smoothly.

\subsubsubsection{Interjection ratio}\!\!\!---
The interjection ratio is the ratio of the frequency of interjections within a text to the total number of words. 
Interjections are words or phrases that express sudden or spontaneous emotion. They make the text appear 
more lively and mirroring of real life. A higher interjection ratio speaks to a higher level of emotion within the text.

\subsubsection{Structural Features}

Structural features deal with the construction and architecture of the text. They provide information on how sentences are formed and how 
paragraphs are organized. They also contribute to the high-level structure of the text. This includes aspects such as sentence length, 
complexity, and the use of lower and upper case letters. Together all of these characteristics influence the readability and 
aesthetics of the text. These features are crucial in order to gain a deeper understanding of how ideas are presented in text. 
They provide a way in which we can understand the organizational preferences of the author.
Next, we introduce the eight structural features that we consider.

\subsubsubsection{Average sentence length}\!\!\!---
Average sentence length is, of course, the average number of words per sentence in a text. Longer sentences often 
contain more complex ideas or multiple thoughts joined together. On the other hand, shorter sentences tend to be 
more concise and focused on one idea. The average sentence length gives us an indication of the complexity 
of the text, and the writing style of the author.

\subsubsubsection{Lowercase letter ratio}\!\!\!---
The lowercase letter ratio is the ratio of the number of lowercase alphabetic characters to the total 
number of alphabetic characters in the paragraph. Texts containing a large value for this ratio indicate 
the adherence of the writer to traditional writing conventions, which state that capitalization should only 
be used to begin sentences or to name locations and persons. A lower ratio could be a sign of unusual structural choice, 
which may be an indication of a human author.

\subsubsubsection{Capital letter ratio}\!\!\!---
The uppercase letter ratio is the ratio of the number of uppercase alphabetic characters to the total number of 
alphabetic characters in the paragraph. A higher ratio could be a sign of unusual structural choice, which may be an 
indication of a human author.

\subsubsubsection{Lexical diversity}\!\!\!---
Lexical diversity refers to the number of unique words used in the text. A higher value indicates the use of a wide range of 
vocabulary within the text. In creative writing, high lexical diversity can contribute to the vividness of the text. This allows 
the writer to capture and convey more complex emotions effectively. In academic writing, a higher degree of lexical diversity 
is often associated with greater sophistication.

\subsubsubsection{Sentence complexity}\!\!\!---
Our sentence complexity score is the average number of clauses per sentence in a given text. It provides information about 
the syntactic complexity of the text by indicating how tightly clauses are packed into sentences. A higher score denotes 
more complex sentence structure, which indicates more complex concepts and a greater degree of information. 
However, this higher complexity can also make the text harder to read.

\subsubsubsection{Burstiness}\!\!\!---
Burstiness is a characteristic that represents to what extent words occur in certain ``bursts'' or clusters, 
rather than in an even distribution throughout the paragraph. Burstiness is common among humans and 
is often a characteristic of specific authors. Burstiness is calculated as the ratio of the variance to the mean 
in the frequency of words occurring within a given text. 

\subsubsubsection{Sentence count}\!\!\!---
Sentence count is the total number of sentences in a paragraph. This can help us gauge whether the chatbot tends
to write longer or shorter sentences, compared to a typical human. For instance, a higher sentence count with shorter 
average sentence length might suggest a style that prioritizes clarity and simplicity.  

\subsubsubsection{Word count}\!\!\!---
Word count is the total number of words in the paragraph. Analyzing the word count can offer insights into the style of the text. 
For example, a higher word count with complex sentence structures may indicate a more detailed text, which is typical  
in scholarly articles. Conversely, a lower word count with simple sentences might be more appropriate for a quick read 
or content aimed at a broader audience.

\subsubsubsection{Stopword ratio}\!\!\!---
The stopword ratio in a text is the proportion of commonly used words that carry minimal lexical content (e.g., ``the,'' ``is,'' ``at''). 
This statistic helps in assessing the density of meaningful content in a given text.

\subsubsubsection{Complex ratio}\!\!\!---
Complex sentences are those that contain one independent clause and at least one dependent clause, 
linked by subordinating conjunctions or relative pronouns. Such sentences tend to express deeper 
relationships and nuances in ideas.

\subsubsection{Semantic Features}

Semantic features deal with the meaning of the text, including the ideas and emotions conveyed by the paragraph. 
This includes the sentiment of the text, figures of speech, literary devices, subjectivity, and objectivity of the text. 
These features showcase the richness of the language used. Next, we introduce the six semantic features
that we consider.

\subsubsubsection{Sentiment polarity}\!\!\!---
Sentiment polarity refers to the sentiment of the text, which can be positive, negative, or neutral. This gives us an 
idea of the mood conveyed by the text and the writer's attitude towards the subject matter. A score between~$\hbox{}-1$ 
and~$\hbox{}+1$ is assigned to each paragraph, where~$\hbox{}-1$ refers to an extremely negative sentiment, 
0 is neutral and~$\hbox{}+1$ is extremely positive.

\subsubsubsection{Subjectivity}\!\!\!---
Subjectivity quantifies what percent of the text is the writer's opinion, feelings, or personal experiences. Human text 
is likely to be more subjective, whereas model-generated text tends to be objective and neutral, eschewing personal 
opinions. A score between~0 and~1 is assigned to each paragraph, with~1 being completely subjective 
and~0 being completely objective.

\subsubsubsection{Homonym frequency}\!\!\!---
Homonyms refer to words that sound or spell the same but have different meanings. Higher homonym frequency 
might indicate a text with more potential ambiguities, which is usually a characteristic of poetic or literary works. 
Such texts often require more contextual interpretation. Conversely, lower homonym frequency could suggest clearer 
and more straightforward text that is intended for a wide audience with varying levels of language proficiency. 

\subsubsubsection{Simile frequency}\!\!\!---
Simile frequency measures the use of the simile figure of speech. Simile refers to the comparison of two different things,
typically by using the words ``as'' or ``like.'' Similes are used to draw parallels while enhancing the language used. 
The strategic use of similes can enrich a narrative, making descriptions more engaging.

\subsubsubsection{Synonym frequency}\!\!\!---
Synonym frequency is the frequency of synonyms used in a paragraph. Synonyms refer to different words that have the same 
meaning. Synonyms help avoid the repetition of words and enhance the diversity of words.

\subsubsubsection{Antonym frequency}\!\!\!---
Antonym frequency is the measure of how often opposites are used in the paragraph. They help enrich the text by adding 
contrast and depth. The use of antonyms can create a more vivid narrative or more persuasive arguments.

\subsubsection{Interaction Features}

Interaction features deal with how the writer engages with the reader. They include conditionals, questions, 
tones, and parts of speech that address dialogue. These feature plays a role in analyzing the strategies used to 
engage the reader and achieve communicative objectives. Next, we introduce the eight interaction features that we consider.

\subsubsubsection{Active passive ratio}\!\!\!---
Active passive ratio is the ratio of the frequency of active to passive voice. Active voice refers to a more direct approach 
to sentences, where the subject of the sentence performs the action described by the verb. The structure of an active voice 
sentence is 
$$
  \mbox{subject} + \mbox{verb} + \mbox{object} 
$$
On the other hand, in passive-voiced sentences, 
the focus of the sentence is the action or object rather than the subject. 
A typical passive voice sentence structure is
$$
  \mbox{object} + \mbox{past participle of verb} + \mbox{subject}
$$
Writing style is greatly affected by the voice used.

\subsubsubsection{Direct to indirect speech ratio}\!\!\!---
Direct speech refers to quoting a speaker without changes. In contrast, indirect speech refers to paraphrasing 
the speaker, without the use of quotes. Direct speech, also called reported speech is used to portray a dialogue. 
Indirect speech is usually used for summarizing existing conversations.

\subsubsubsection{Conditional sentence ratio}\!\!\!---
Conditional sentence ratio is the ratio of the number of conditional sentences to the total number of sentences in a text. 
Conditional sentences refer to statements that include a hypothetical situation as an outcome using 
connectors such as ``if'', ``unless'', and so on.

\subsubsubsection{Negation ratio}\!\!\!---
Negation ratio is the ratio of sentences containing negations to the total number of sentences in a text. 
Negations usually refer to the use of words such as ``not'', ``no'', or ``never'', which effectively invert 
the meaning of the sentence. They are used to express contradictions or disagreements within the text.

\subsubsubsection{Question ratio}\!\!\!---
The question ratio is the ratio of the number of questions in the text to total number of sentences. In the context of our dataset,
questions will generally be more rhetorical and meant to get the reader to ponder over the content, and is
a common human author strategy.

\subsubsubsection{Exclamatory sentence ratio}\!\!\!---
The exclamatory sentence ratio is the ratio of the number of exclamatory sentences to total number of sentences. 
Exclamatory sentences express strong feelings, reactions, surprise, excitement, or other intense emotions,
and typically end with an exclamation point. Such sentences are often used for dramatic effect. 

\subsubsubsection{Imperative mood ratio}\!\!\!---
Imperative mood ratio is the ratio of sentences containing imperative mood to the total number of sentences. 
An imperative sentence demands or requires that an action be taken, and
texts with a higher ratio of imperative mood are usually instructive in nature.

\subsubsubsection{Subjunctive mood ratio}\!\!\!---
The subjunctive mood is used to express wishes, hypotheticals, or contrary scenarios. 
A higher ratio of subjunctive mood indicates more speculative context, and is often found in 
literary works, opinion pieces, or discussions involving scenarios that are not grounded in reality.

\section{Data Exploration}\label{data exploration}

Before discussing the ML models cosidered, we explore what insights can be 
gathered directly from the dataset. This includes an analysis of the distribution of the data across 
the various features introduced in the previous section. 

\subsection{Target Variable}

The target variable that we attempt to predict is whether a given piece of text is ``human'' or ``GPT''. 
For training, the dataset consists of the text from wikiHow articles and corresponding GPT-generated text. 
Therefore, in terms of the number of paragraphs, the dataset is balanced.

\subsection{Paragraph Length}

\begin{figure}[!htb]
    \centering
    \includegraphics[width=0.8\textwidth]{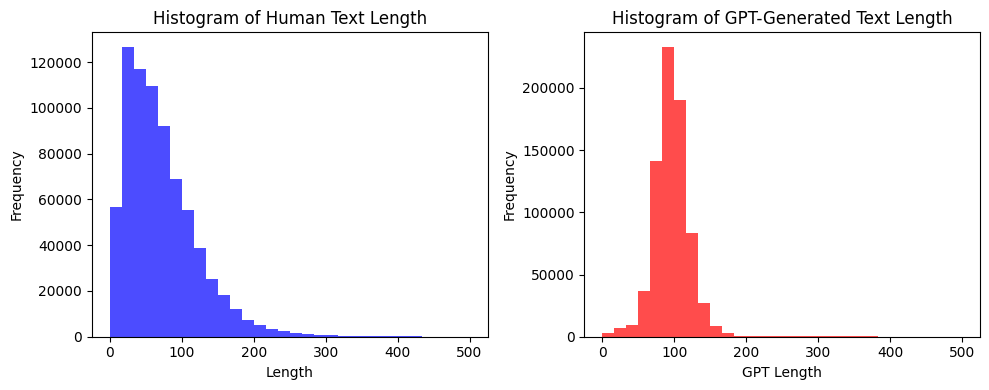}
    \caption{Distribution of number of words in paragraphs}\label{fig:length_distribution}
\end{figure}

While creating the GPT generated counterpart of the human written data, 
we passed the length of the paragraph to the prompt, with the goal of generating
reasonably consistent lengths in the corresponding GPT generated data. 
Figure~\ref{fig:length_distribution} reflects this distribution of the target variable.

From Figure~\ref{fig:length_distribution} it is clear that the distribution of words in human as compared to GPT is significantly 
different, despite passing the length of human written paragraphs to the GPT model in the prompt. This indicates that the model 
is not able to strictly stay within the specified word length. One reason for this is that the GPT model is based on ``tokens'' instead of 
words, and one word can contain anywhere from one to three tokens, on average. Hence, the model is not able to adhere well 
to word limits. Figure~\ref{fig:length_distribution} indicates that GPT tends to be more wordy or chatty,
as compared to humans.

\subsection{Feature Analysis}

This section outlines the insights gained from visualizing the target variable with individual features. 
This can help us understand how these features affect the predictor variable. The graphs from which 
these insights were drawn can be found in Appendix~\ref{appendix_b}.

\subsubsection{Linguistic Features}

By examining the linguistic patterns in human and GPT-generated text, we can observe clear variations in different 
parts of speech. For instance, humans typically use verbs more often, and in addition, have a higher mean verb ratio. 
They also use a wider variety of verbs compared to GPT. 
Similarly, human texts have a wider range of prepositions. This suggests that GPT-generated texts have a more
structured and limited approach to the usage of prepositions.

Another interesting difference between human vs GPT-generated texts is the usage of adjectives. 
GPT-generated texts tend to use more adjectives and in turn more descriptive language. This might 
be done in an effort to enhance the quality and add depth to the text. In contrast, the usage of pronouns is 
higher in humans. This shows that humans generally prefer a more personal and interactive style of writing. 

Despite the use of interjections being limited in both human and GPT text, their use in human text is 
relatively more frequent. This indicates that human texts are more spontaneous, and that this is one characteristic 
that GPT does not attempt to imitate. This is a subtle difference between the two classes.

\subsubsection{Structural Features}

Structural features offer more insights into the nuanced differences between human-generated 
and GPT-generated texts. Sentences produced by GPT are typically longer than sentences produced by humans. 
Human-written sentences have a wider range of sentence lengths, in contrast to the consistently longer sentences 
of GPT. Additionally, the use of capitalization is also higher in humans, mostly due to emphasis and stylistic choices.

In addition to sentence variety and capitalization, human writing also has a greater diversity of words, indicating that
humans tend to use more creative language. This is not surprising, as a trained chatbot tends to select
the best fitting word (or words), whereas humans often prefer more variation to make the text more
interesting to read. 

Sentence composition is another characteristic that differs in human as compared to GPT generated text. 
GPT is known for generating complex sentences with multiple clauses. Human writing, on the other hand, 
demonstrates a wider range of complexity; some authors favor straightforward sentence structures, 
while others choose far more complexity. 

Burstiness is an attribute that describes the repeated occurrence of certain words or phrases in ``bursts''. 
Human writing has a higher mean and variance in burstiness, while 
GPT text is generally less erratic, highlighting a more consistent approach to text generation. 

Due to longer sentences of GPT text, it also tends to use more words overall. Hence, GPT-generated texts 
have a higher level of verbosity or wordiness, and the wordiness of AI writing occasionally results in redundancy. 
Humans, on the other hand, write succinct and to-the-point sentences focusing on clarity. 

\subsubsection{Semantic Features}

Humans tend to include personal thoughts, opinions, and emotional range in their writing. 
This leads to varied tones and perspectives. In contrast, GPT texts are consistently neutral or positive. 
This is likely the result of an algorithmic bias that produces less divisive content to avoid controversy. 
In addition, GPT tends to stick to facts, without expressing opinions. This might be due to the fact that 
chatbots are intended to be informative and subjective, without introducing their own bias.

Another interesting characteristic of GPT text is the frequent use of homonyms. It is unclear
why ChatGPT tends to use more homonyms than humans.

Conversely, GPT tends to use more similes than humans. This might be because it is trained to enhance the readability 
of text and make it more likely to be understood. Additionally, GPT also uses more synonyms than humans while, in contrast, 
GPT uses fewer antonyms, preferring non-contrasting language. This might again stem from the fact that it is trained to make 
the content less divisive and polarizing. Humans, not bound by this restriction, tend to use a larger number of antonyms.

\subsubsection{Interaction Features}

Interaction features refer to how the text interacts with the reader, such as the ``voice'' of the text.  
Human text has a higher median of active to passive ratio with a wider variation, which suggests that human 
authors use a variety of stylistic choices when picking the voice of the sentences. ChatGPT, on the other hand, 
utilizes an active voice more frequently, thus making the text direct, straightforward, and easier to understand.

Text written by humans has a larger mean conditional sentence ratio than that generated by GPT. 
Humans tend to use conditionals (e.g., ``if'') more often to discuss possibilities and hypotheticals. 
Human-generated text also has a larger mean negation ratio than GPT, indicating a higher frequency 
of negative constructs in human language. GPT uses fewer negations which, again, serves to keep the 
text neutral and less polarizing.

Humans use questions more frequently than GPT as evidenced by the higher mean question ratio. 
Questions are often employed to engage with the reader. This suggests a more interactive approach in 
human writing compared to GPT's generally more informative style.

Exclamatory sentences, which express strong feelings and reactions, appear at a higher ratio in human texts. 
This contrasts with the typically more subdued tone of GPT-generated texts.

Lastly, the mean subjunctive mood ratio is larger in texts created by humans. This indicates a greater tendency for 
humans to discuss hypothetical scenarios, which requires thinking beyond the immediate. 
This is less frequently observed in GPT text.

\subsection{Correlation Analysis}

\begin{figure}[!htb]
    \centering
\begin{tikzpicture}[scale=0.5]
    \begin{axis}[
        width=18cm,
        height=18cm,
	colormap={redblue}{color=(blue) rgb255=(255, 0, 13)},
        xticklabels={
lexical\_diversity,
homonym\_frequency,
synonym\_frequency,
verb\_ratio,
sentence\_count,
verb\_present\_tense\_frequency,
lowercase\_letter\_ratio,
auxilary\_verb\_frequency,
capital\_letter\_ratio,
word\_count
        },
        xtick={0,...,9},
        xtick style={draw=none},
	xticklabel style={anchor=east,rotate=60,yshift=-5pt,font=\tt,scale=1.5},
        yticklabels={
lexical\_diversity,
homonym\_frequency,
synonym\_frequency,
verb\_ratio,
sentence\_count,
verb\_present\_tense\_frequency,
lowercase\_letter\_ratio,
auxilary\_verb\_frequency,
capital\_letter\_ratio,
word\_count
        },
        ytick={0,...,9},
        ytick style={draw=none},
        enlargelimits=false,
        yticklabel style={font=\tt,scale=1.5},
        colorbar,
        colorbar style={
            ytick={-1.0,-0.75,-0.5,-0.25,0.0,0.25,0.5,0.75,1.0},
            yticklabels={-1.0,-0.75,-0.5,-0.25,0.0,0.25,0.5,0.75,1.0},
            yticklabel={\pgfmathprintnumber\tick},
            yticklabel style={
            		scale=1.5,
            		/pgf/number format/fixed,
			/pgf/number format/fixed zerofill,
			/pgf/number format/1000 sep={},
			/pgf/number format/precision=2}
        },
        point meta min=-1.0,
        point meta max=1.0,
        nodes near coords={\pgfmathprintnumber\pgfplotspointmeta},
        nodes near coords black white/.style={
            small value/.style={
                yshift=8pt,
                text=white,
                /pgf/number format/fixed,
                /pgf/number format/1000 sep={},
                /pgf/number format/precision=3,
                /pgf/number format/zerofill=true,
                scale=1.0,
            },
            large value/.style={
                yshift=-8pt,
                text=white,
                /pgf/number format/fixed,
                /pgf/number format/1000 sep={},
                /pgf/number format/precision=3,
                /pgf/number format/zerofill=true,
                scale=1.0,
            },
            every node near coord/.style={
                check for zero/.code={
                    \pgfmathfloatifflags{\pgfplotspointmeta}{0}{
                        \pgfkeys{/tikz/coordinate}
                    }{
                        \begingroup
                        \pgfkeys{/pgf/fpu}
                        \pgfmathparse{\pgfplotspointmeta<#1}
                        \global\let\result=\pgfmathresult
                        \endgroup
                        %
                        %
                        \pgfmathfloatcreate{1}{1.0}{0}
                        \let\ONE=\pgfmathresult
                        \ifx\result\ONE
                            \pgfkeysalso{/pgfplots/small value}
                        \else
                            \pgfkeysalso{/pgfplots/large value}
                        \fi
                    }
                },
                check for zero,
            },
        },
        nodes near coords black white=0.0,
    ]
        \addplot[
            matrix plot,
            mesh/cols=10,
            point meta=explicit,draw=gray
        ] table [meta=C] {
            x y C
0 0 1.00
1 0 -0.16
2 0 0.63
3 0 0.25
4 0 0.80
5 0 0.87
6 0 -0.083
7 0 0.98
8 0 0.94
9 0 -0.70
0 1 -0.16
1 1 1.00
2 1 -0.16
3 1 -0.63
4 1 -0.25
5 1 -0.07
6 1 -0.10
7 1 -0.21
8 1 -0.27
9 1 0.15
0 2 0.63
1 2 -0.16
2 2 1.00
3 2 0.088
4 2 0.59
5 2 0.55
6 2 0.25
7 2 0.62
8 2 0.55
9 2 -0.43
0 3 0.25
1 3 -0.63
2 3 0.088
3 3 1.00
4 3 0.35
5 3 0.13
6 3 -0.16
7 3 0.33
8 3 0.42
9 3 -0.17
0 4 0.80
1 4 -0.25
2 4 0.59
3 4 0.35
4 4 1.00
5 4 0.68
6 4 -0.044
7 4 0.80
8 4 0.79
9 4 -0.53
0 5 0.87
1 5 -0.07
2 5 0.55
3 5 0.13
4 5 0.68
5 5 1.00
6 5 -0.027
7 5 0.85
8 5 0.81
9 5 -0.59
0 6 -0.083
1 6 -0.10
2 6 0.25
3 6 -0.16
4 6 -0.044
5 6 -0.027
6 6 1.00
7 6 -0.11
8 6 -0.13
9 6 0.004
0 7 0.98
1 7 -0.21
2 7 0.62
3 7 0.33
4 7 0.80
5 7 0.85
6 7 -0.11
7 7 1.00
8 7 0.96
9 7 -0.68
0 8 0.94
1 8 -0.27
2 8 0.55
3 8 0.42
4 8 0.79
5 8 0.81
6 8 -0.13
7 8 0.96
8 8 1.00
9 8 -0.62
0 9 -0.7
1 9 0.15
2 9 -0.43
3 9 -0.17
4 9 -0.53
5 9 -0.59
6 9 0.004
7 9 -0.68
8 9 -0.62
9 9 1.00
         };
    \end{axis}
\end{tikzpicture}
    \caption{Correlation heatmap}\label{fig:correlation}
\end{figure}
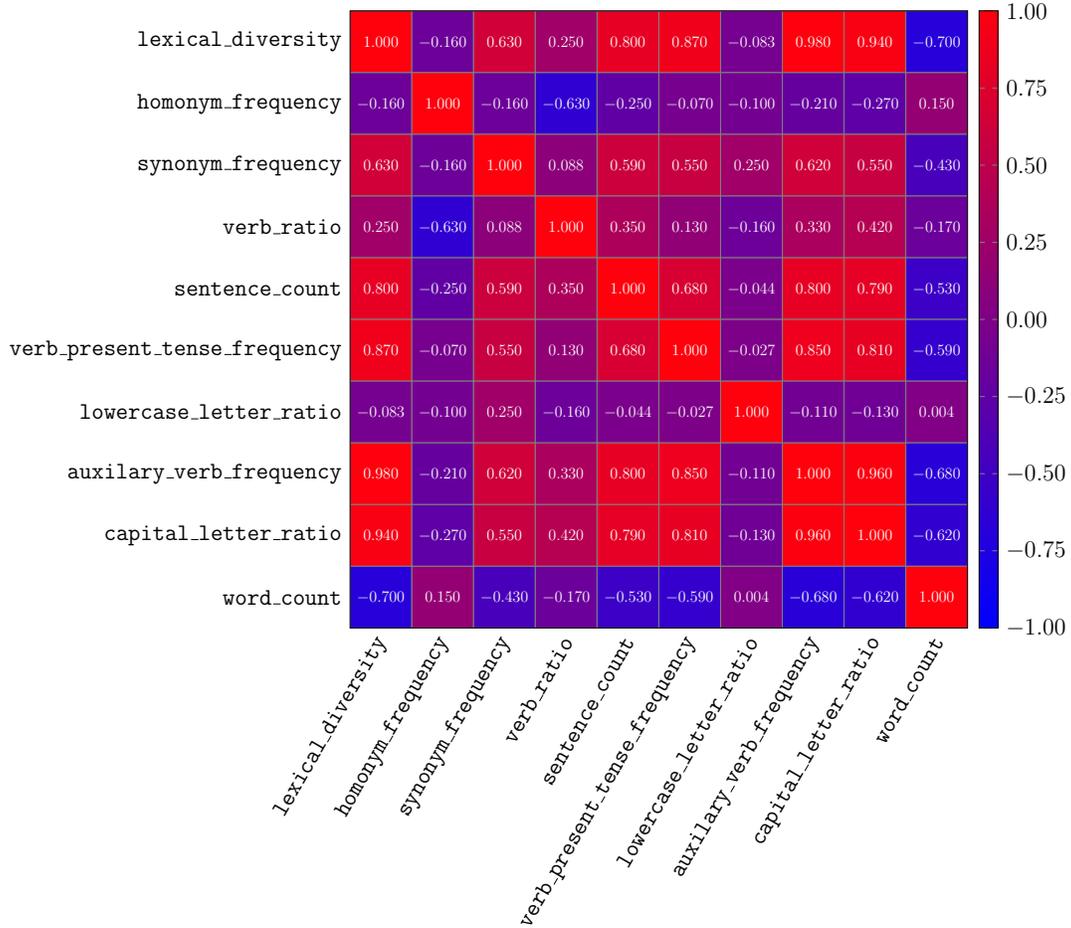

Figure~\ref{fig:correlation} provides a heatmap of the correlations between pairs of features. For example, we
observe that \texttt{auxiliary\_verb\_frequency} and \texttt{lexical\_diversity}
are highly correlated. As another example, we note 
that \texttt{homonym\_frequency} is not highly correlated with
any of the other features.

\section{Implementation}\label{implementation}

This section describes the process through which we trained ML and DL models to distinguish between 
human-generated and GPT-generated text. We provide details about our feature analysis approach, 
which involves crafting and selecting informative features from the text data. We also discuss our embeddings approach, 
where text is converted into numerical representations that capture deeper semantic meanings.

\subsection{Data Preprocessing}

Generally, preparing the data to a point where models can be trained on it is one of the most important steps 
in ML. This step typically involves cleaning the dataset, removing unnecessary or redundant features, 
analyzing correlation, and selecting the appropriate features for training. 
However, since data generation is part of this research, most preprocessing tasks have already been addressed during the 
dataset creation phase. Consequently, minimal preprocessing is necessary before proceeding with model training. 
This section describes the minimal preprocessing required by the models.
In fact, the only significant processing that we need to perform
is to split the data into training and test sets. In all of our experiments, the dataset 
is split at an~80-20 ratio for training and testing, respectively. Note that due to the
number of experiments and the size of the dataset, we do not perform cross-validation.

\subsection{Feature Importance}\label{sect:faa}

In our feature analysis approach, we generate features from the raw data, and these features are then used for training binary 
classification models. The feature generation process involves extracting characteristics from the text, 
including syntactical features, semantic features, structural features, and interaction features, as discussed in
Section~\ref{sect:FA}, above. These features aim to capture subtle differences between the structured nature of 
machine-generated text and the more variable style of human writing.

Following feature extraction, we employ various feature selection techniques such as 
Principal Component Analysis (PCA), Linear Discriminant Analysis (LDA), feature importance charts, etc. 
This is done to identify the most important features for distinguishing between human and GPT-generated writing. 
This step helps in reducing dimensionality and improving model efficiency by eliminating redundant or less informative features.

The selected features are then utilized to train several types of binary classification models. We explore a range of 
models---from simple Logistic Regression to more complex neural networks, such as LSTMs. Finally, the performance 
of each model is evaluated (based on accuracy), and the models are compared.

Once we have the important features, we modify features to see how such changing affect model performance. 
This helps us identify non-human-like aspects of the chatbot.

\subsubsection{Feature Selection Techniques}

Feature selection is a critical process in machine learning that involves identifying and selecting the features that 
have the most significant impact on the predictor variable. This process is essential because it directly influences
the performance of the machine learning models. By focusing on the most relevant features, we can enhance the 
model accuracy, reduce overfitting, and decrease the computational cost associated with training, and less data 
would need to be collected when the model is used for classification in practice. 

For the classification problem at hand, the extracted features are all assumed to be likely to impact the predictor variable. 
However, this might not be true for all the features, as some might be of minimal value. Moreover, some features might be 
correlated, thus adding little or no new information. Therefore, it is necessary to identify the subset of input variables that 
are most predictive of the desired outcome. By eliminating unnecessary and redundant features, the model can concentrate 
its learning on the aspects of the data that are most distinguishing. This helps in improving both the training efficiency and 
the generalizability of the model.

There are many feature selection techniques. For feature reduction, we consider Principal Component Analysis (PCA), 
Linear Discriminant Analysis (LDA), as well as feature selection based on a Random Forest classifier,
and Lasso feature importance. Next, we briefly describe each of these feature reduction techniques.

\subsubsubsection{Principal Component Analysis}\!\!\!---
PCA is a widely used technique in ML for dimensionality reduction. It simplifies data with a large number of dimensions 
while retaining statistically significant aspects of the original data. PCA works by identifying the directions along which the 
variances within the data is maximized. These directions, known as principal components, are linear combinations of the 
original features. Moreover, the principal components are orthogonal, ensuring that they are uncorrelated~\cite{stampML}.

\subsubsubsection{Linear Discriminant Analysis}\!\!\!---
Like PCA, Linear Discriminant Analysis is also a dimensionality reduction technique that is used to find a linear combination of features. 
However, unlike PCA, LDA explicitly considers class labels to identify the principal components that maximize the separation between 
multiple classes~\cite{stampML}. 

\subsubsubsection{Random Forest}\!\!\!---
Random Forest models are often used to evaluate and rank the importance of various features in a dataset.
The significance of each feature is evaluated based on its impact on the Random Forest model accuracy. 
This significance is calculated by observing how much the accuracy of the model decreases when the values 
of that feature are randomly shuffled while keeping all other feature values constant. This shuffling changes 
the structure that the feature brings to the model, thus highlighting its influence on the model performance.

\begin{figure}[!htb]
    \centering
    \includegraphics[width=0.6\textwidth, height=0.28\textheight]{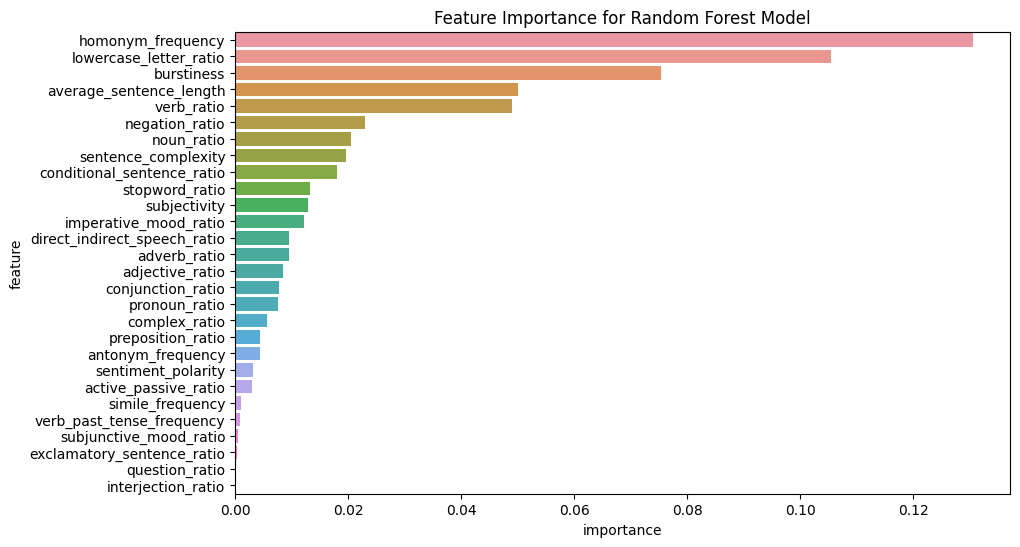}
    \caption{Random Forest feature importances}\label{fig:rf_feature_importances}
\end{figure}

Figure~\ref{fig:rf_feature_importances} gives the random forest feature importance for our dataset. 
For performing feature selection, we identify the top~15 most important features and discard the remaining features. 

\subsubsubsection{Lasso}\!\!\!---
Least Absolute Shrinkage and Selection Operator (Lasso) is a modification of linear regression 
that incorporates a regularization term in the loss function. This term, known as an~L1 penalty, 
is directly proportional to the absolute value of the coefficient magnitudes. 

The Lasso regression technique can be used to perform both feature selection and regularization. 
By introducing the~L1 penalty term, Lasso converts the coefficients of less important features to zero. 
This aspect of Lasso is useful in feature selection as it automatically reduces the number of features 
by setting some coefficients to zero. The greater the value of the penalty term, the more coefficients are set 
to zero. The remaining features (i.e., those with non-zero coefficients) are considered significant.

\begin{figure}[!htb]
    \centering
    \includegraphics[width=0.6\textwidth, height=0.28\textheight]{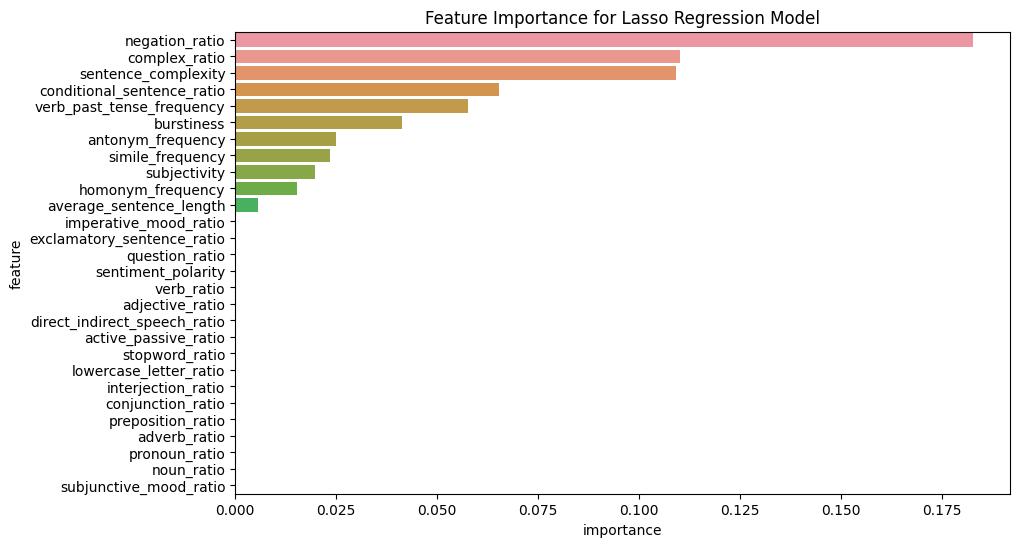}
    \caption{Lasso feature importances}\label{fig:lasso_feature_importances}
\end{figure}

Figure~\ref{fig:lasso_feature_importances} illustrates the features coefficients. 
We observe that~11 of the features have non-zero coefficients, and hence we use
these features in our Lasso experiments.

To assess the impact of these various feature selection techniques on model performance, 
we conduct numerous experiments. As outlined above, the feature selection techniques that we have 
evaluated are the following.
\begin{itemize}
    \item No feature selection
    \item Principal Component Analysis (PCA)
    \item Linear Discriminant Analysis (LDA)
    \item Features selected by Random Forest
    \item Features selected by Lasso
\end{itemize}
For each of these feature selection methods, six distinct models have been trained to compare their effectiveness. 
These models are the following.
\begin{itemize}
    \item Logistic Regression
    \item Random Forest (RF)
    \item XGBoost
    \item Multilayer Perceptron (MLP)
    \item Deep Neural Network (DNN)
    \item Long Short Term Memory (LSTM)
\end{itemize}
This comprehensive approach allows us to thoroughly evaluate how the various feature selection strategies 
influence the predictive accuracy and of a wide variety of machine learning and deep learning models. 
Note that each of the six models considered was discussed in Section~\ref{background:models}, above.

\subsubsection{Similar-Length Dataset}\label{new_data}

Figure~\ref{fig:length_distribution} indicates that the distribution of paragraph lengths is 
different in human-generated text, as compared to GPT-generated text. 
We normalize most features based on 
the total number of words, but this uneven distribution of words might skew some features. 
For example, from Figure~\ref{fig:rf_feature_importances} we see that 
\texttt{homonym\_frequency} is the most important feature, according to a Random Forest model.

\begin{figure}[!htb]
    \centering
    \includegraphics[width=0.7\textwidth, height=0.33\textheight]{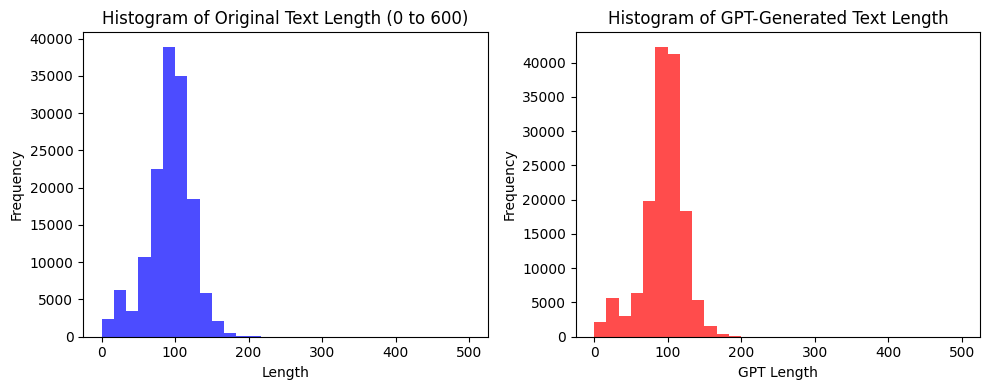}
    \caption{Paragraph length distribution for similar-length subset}\label{fig:equal_length_distribution}
\end{figure}

As an experiment, we wanted to consider the influence of text length on the trained models, that is, how the
various models perform over text with similar length distributions in each class. To test this case, 
we selected a subset of our data for which the absolute difference in length between the human-generated
paragraph and the corresponding chatbot-generated text is less than~15 words. We refer to this as the
similar length-length subset, and it consists of~292,604 paragraphs.
Figure~\ref{fig:equal_length_distribution} shows the distribution of the number of words in
each paragraph within this similar-length subset.

\begin{figure}[!htb]
    \centering
    \includegraphics[width=0.7\textwidth, height=0.33\textheight]{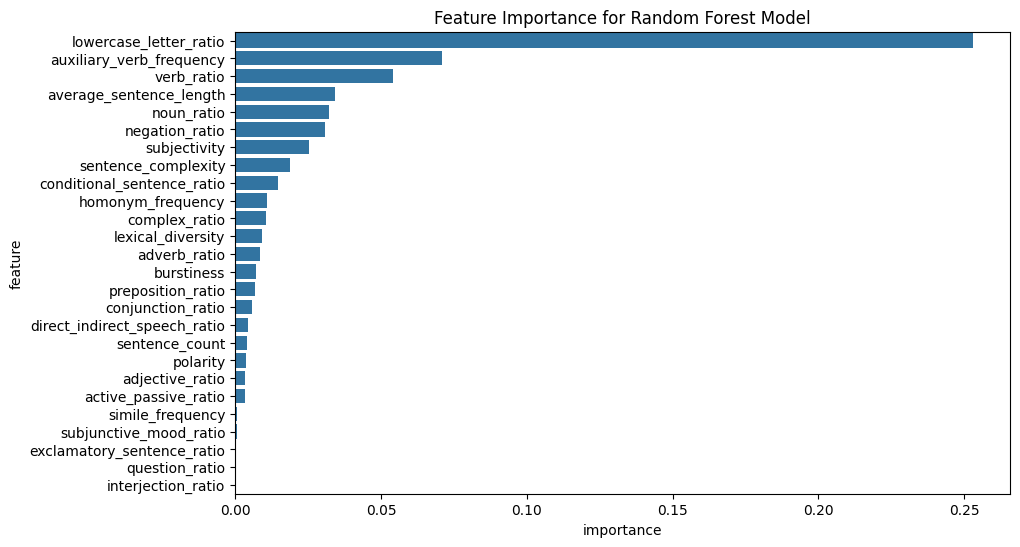}
    \caption{Random Forest feature importance chart for similar-length data}\label{fig:rf_feature_importance_small}
\end{figure}

\begin{figure}[!htb]
    \centering
    \includegraphics[width=0.7\textwidth, height=0.33\textheight]{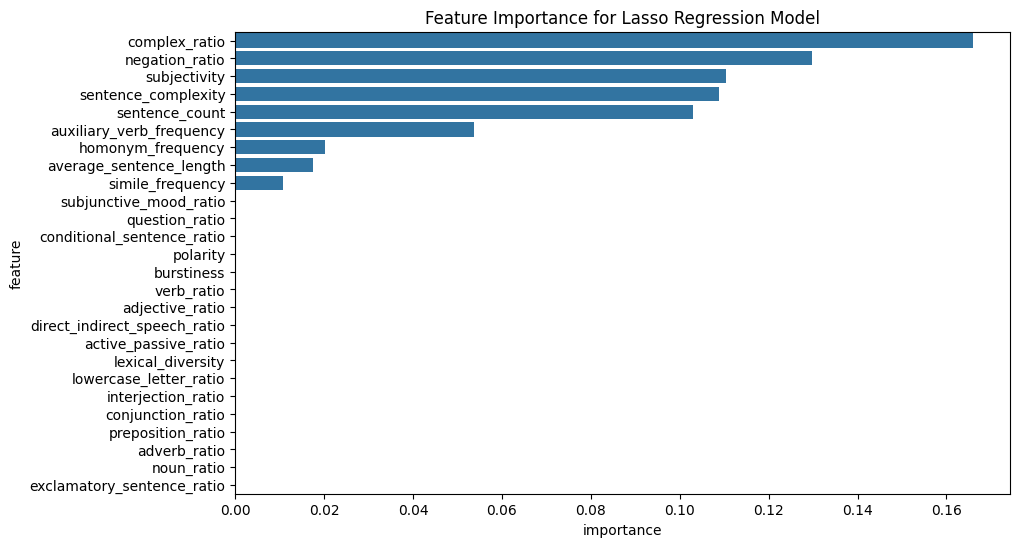}
    \caption{Lasso feature importance chart for similar-length data}\label{lasso_feature_importance_small}
\end{figure}

The same 
feature extraction, data preprocessing, and feature selection steps outlined in Section~\ref{sect:faa} 
have been carried out on our similar-length subset. Figures~\ref{fig:rf_feature_importance_small} 
and~\ref{lasso_feature_importance_small} show the Random Forest and Lasso feature importances
respectively. 
From Figures~\ref{fig:rf_feature_importance_small}, we observe that the importance
of \texttt{homonym\_frequency} has been reduced.

\subsection{Embeddings Approach Implementation}

Our embedding approach involves using sophisticated algorithms to convert text into numerical vector representations.
These embeddings are then analyzed by machine learning 
models to detect differences in language patterns.

Vector embeddings are designed to capture semantic and syntactic nuances of 
the language. This is done by analyzing large texts and learning representations where related words have related 
encodings. These models provide dense and informative representations that capture context and meaning effectively.
The embeddings techniques that we consider are the following: TFIDF, Word2Vec, GloVe, and BERT.
Each of these word embedding techniques was introduced in Section~\ref{sect:embed}, above.
As with our feature selection experiments, for each of these embedding techniques, 
the same six learning models have been trained to compare their effectiveness. 
These models are the following: LR, RF, XGBoost, MLP, DNN, LSTM.
This comprehensive approach allows us to thoroughly evaluate how these various embedding techniques
affect the accuracy of a wide variety of learning models. 
Again, each of the six models considered was introduced in Section~\ref{background:models}, above.

\section{Experimental Results}\label{results}

In this section, we present and analyze our experimental results. We first consider the
results of our feature analysis experiments---for both the case where we use all of the
data, and for the case where we use a subset of the data for which corresponding human and
chatbot written paragraphs are restricted to be nearly the same length. Then we present
the results of our embeddings-based experiments. Finally, we summarize all
of our experimental results.

\subsection{Feature Analysis Experiments}

For the experiments discussed in this section, 
we train models based on the features discussed in Section~\ref{sect:faa}.
As mentioned above, we consider two cases---first, where we use all
of our data, and a second set where we restrict our attention to a subset
where the lengths of the human-generated and corresponding chatbot-generated
paragraphs are nearly the same.

\subsubsection{All of the Data}

\begin{figure}[!htb]
    \centering
    \begin{tikzpicture}[scale=0.8, every node/.style={scale=0.95}]
\pgfkeys{/pgf/number format/.cd,1000 sep={}}
\begin{axis}[
        width  = 0.95*\textwidth,
        height = 7.5cm,
        ymin=0.0,ymax=1.2,
        ytick={0.0, 0.2, 0.4, 0.6, 0.8, 1.0},
        major x tick style = transparent,
        ybar=5*\pgflinewidth,
        bar width=8.0pt,
        ylabel = {Accuracy},
        symbolic x coords={LR, RF, XGBoost, MLP, DNN, LSTM},
        xticklabels={LR, RF, XGBoost, MLP, DNN, LSTM},
	y tick label style={
    		/pgf/number format/.cd,
   		fixed,
   		fixed zerofill,
    		precision=2},
        xtick = data,
        x tick label style={
		font=\small,
		},
        nodes near coords,
        every node near coord/.append style={rotate=90, scale=0.75,
        								   anchor=west, 
								   /pgf/number format/.cd,
								   fixed,
								   fixed zerofill,
								   precision=4},
        enlarge x limits=0.105,
        legend cell align=left,
        legend pos=south east,
]
\addplot [fill=blue,opacity=1.00]
coordinates {
(LR, 0.8647)
(RF, 0.9621)
(XGBoost, 0.9643)
(MLP, 0.9602)
(DNN, 0.9617)
(LSTM, 0.9593)
};
\addlegendentry{All features}
\addplot [fill=red,opacity=1.00]
coordinates {
(LR, 0.8713)
(RF, 0.9520)
(XGBoost, 0.9537)
(MLP, 0.9653)
(DNN, 0.9545)
(LSTM, 0.9523)
};
\addlegendentry{PCA}
\addplot [fill=green,opacity=1.00]
coordinates {
(LR, 0.9243)
(RF, 0.9545)
(XGBoost, 0.9236)
(MLP, 0.9234)
(DNN, 0.9552)
(LSTM, 0.9207)
};
\addlegendentry{LDA}
\addplot [fill=yellow,opacity=1.00]
coordinates {
(LR, 0.8501)
(RF, 0.9473)
(XGBoost, 0.9585)
(MLP, 0.9569)
(DNN, 0.9415)
(LSTM, 0.9467)
};
\addlegendentry{RF}
\addplot [fill=orange,opacity=1.00]
coordinates {
(LR, 0.8570)
(RF, 0.9231)
(XGBoost, 0.9158)
(MLP, 0.9123)
(DNN, 0.9078)
(LSTM, 0.9092)
};
\addlegendentry{Lasso}
\end{axis}
\end{tikzpicture}
    \caption{Accuracy of ML models with various feature selection methods}\label{fig:bar_ML}
\end{figure}
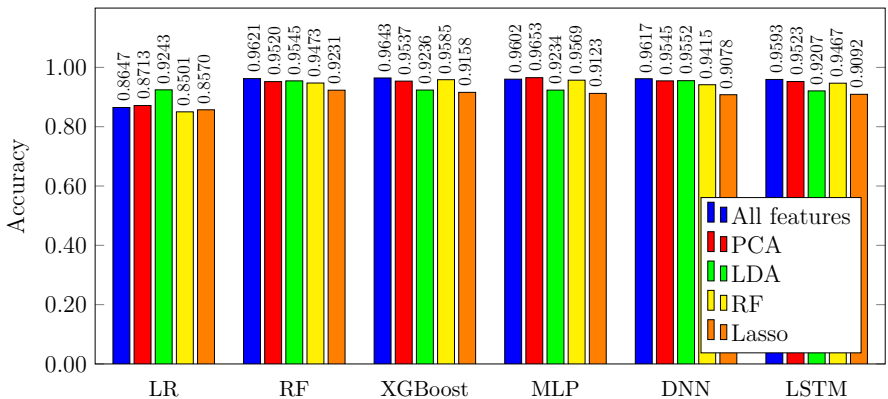

Figure~\ref{fig:bar_ML} shows the accuracies obtained using our feature analysis approach.
Interestingly, feature reduction techniques improve the results for LR, but for the other models considered,
training on all features performs at least as well as training the same model using any of
the feature reduction techniques.


\subsubsection{Similar-Length Experiments}

This section gives the results obtained the similar-length subset 
discussed in Section~\ref{new_data}.
That is, we only consider the subset of our data that is more
balanced with respect to paragraph length.

\begin{figure}[!htb]
    \centering
    \begin{tikzpicture}[scale=0.8, every node/.style={scale=0.95}]
\pgfkeys{/pgf/number format/.cd,1000 sep={}}
\begin{axis}[
        width  = 0.95*\textwidth,
        height = 7.5cm,
        ymin=0.0,ymax=1.2,
        ytick={0.0, 0.2, 0.4, 0.6, 0.8, 1.0},
        major x tick style = transparent,
        ybar=5*\pgflinewidth,
        bar width=8.0pt,
        ylabel = {Accuracy},
        symbolic x coords={LR, RF, XGBoost, MLP, DNN, LSTM},
        xticklabels={LR, RF, XGBoost, MLP, DNN, LSTM},
	y tick label style={
    		/pgf/number format/.cd,
   		fixed,
   		fixed zerofill,
    		precision=2},
        xtick = data,
        x tick label style={
		font=\small,
		},
        nodes near coords,
        every node near coord/.append style={rotate=90, scale=0.75,
        								   anchor=west,
								   /pgf/number format/.cd,
								   fixed,
								   fixed zerofill,
								   precision=4},
        enlarge x limits=0.105,
        legend cell align=left,
        legend pos=south east,
]
\addplot [fill=blue,opacity=1.00]
coordinates {
(LR, 0.8447)
(RF, 0.9234)
(XGBoost, 0.9467)
(MLP, 0.9467)
(DNN, 0.9333)
(LSTM, 0.9352)
};
\addlegendentry{All features}
\addplot [fill=red,opacity=1.00]
coordinates {
(LR, 1.000)
(RF, 0.9999)
(XGBoost, 1.0000)
(MLP, 1.0000)
(DNN, 0.9994)
(LSTM, 0.9996)
};
\addlegendentry{PCA}
\addplot [fill=green,opacity=1.00]
coordinates {
(LR, 0.9173)
(RF, 0.8678)
(XGBoost, 0.9198)
(MLP, 0.9149)
(DNN, 0.9082)
(LSTM, 0.9078)
};
\addlegendentry{LDA}
\addplot [fill=yellow,opacity=1.00]
coordinates {
(LR, 0.8357)
(RF, 0.9279)
(XGBoost, 0.9257)
(MLP, 0.9298)
(DNN, 0.9207)
(LSTM, 0.9224)
};
\addlegendentry{RF}
\addplot [fill=orange,opacity=1.00]
coordinates {
(LR, 0.8264)
(RF, 0.8654)
(XGBoost, 0.8573)
(MLP, 0.8521)
(DNN, 0.8536)
(LSTM, 0.8475)
};
\addlegendentry{Lasso}
\end{axis}
\end{tikzpicture}
    \caption{Accuracy of ML models for similar-length data}\label{fig:bar_SL}
\end{figure}
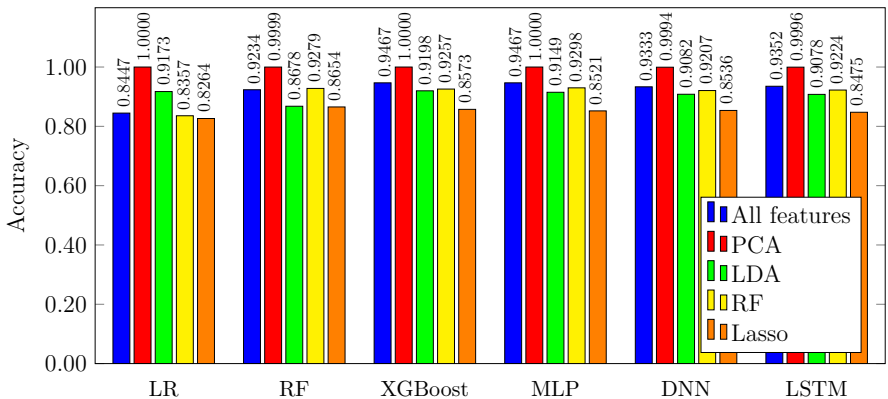

Figure~\ref{fig:bar_SL} gives our results for this similar-length dataset.
In this case, PCA is clearly the best feature reduction approach, with all models using PCA features
having at least~0.99 accuracy. 
Curiously, training models on all features outperforms all of the feature reduction techniques 
(LDA, RF, Lasso) for each of the models considered.


\subsection{Embeddings Approach}

In this approach, each document is converted into a single fixed-length vector by averaging the word embeddings
for each word in a document. This results in fixed-length vectors of length~100 for each paragraph.

\begin{figure}[!htb]
    \centering
    \begin{tikzpicture}[scale=0.8, every node/.style={scale=0.95}]
\pgfkeys{/pgf/number format/.cd,1000 sep={}}
\begin{axis}[
        width  = 0.8*\textwidth,
        height = 7.5cm,
        ymin=0.0,ymax=1.2,
        ytick={0.0, 0.2, 0.4, 0.6, 0.8, 1.0},
        major x tick style = transparent,
        ybar=5*\pgflinewidth,
        bar width=8.0pt,
        ylabel = {Accuracy},
        symbolic x coords={LR, RF, XGBoost, MLP, DNN, LSTM},
        xticklabels={LR, RF, XGBoost, MLP, DNN, LSTM},
	y tick label style={
    		/pgf/number format/.cd,
   		fixed,
   		fixed zerofill,
    		precision=2},
        xtick = data,
        x tick label style={
		font=\small,
		},
        nodes near coords,
        every node near coord/.append style={rotate=90, scale=0.75,
        								   anchor=west,
								   /pgf/number format/.cd,
								   fixed,
								   fixed zerofill,
								   precision=4},
        enlarge x limits=0.10,
        legend cell align=left,
        legend pos=south east,
]
\addplot [fill=blue,opacity=1.00]
coordinates {
(LR, 0.9839)
(RF, 0.9820)
(XGBoost, 0.9801)
(MLP, 0.9876)
(DNN, 0.9786)
(LSTM, 0.9723)
};
\addlegendentry{TFIDF}
\addplot [fill=red,opacity=1.00]
coordinates {
(LR, 0.9154)
(RF, 0.9424)
(XGBoost, 0.9557)
(MLP, 0.9790)
(DNN, 0.9780)
(LSTM, 0.9765)
};
\addlegendentry{Word2Vec}
\addplot [fill=green,opacity=1.00]
coordinates {
(LR, 0.9460)
(RF, 0.9548)
(XGBoost, 0.9673)
(MLP, 0.9825)
(DNN, 0.9775)
(LSTM, 0.9780)
};
\addlegendentry{GloVe}
\addplot [fill=yellow,opacity=1.00]
coordinates {
(LR, 0.9982)
(RF, 0.9673)
(XGBoost, 0.9800)
(MLP, 0.9757)
(DNN, 0.9863)
(LSTM, 0.9862)
};
\addlegendentry{BERT}
\end{axis}
\end{tikzpicture}
    \caption{Accuracy of ML models with various embedding techniques}\label{fig:bar_embed}
\end{figure}
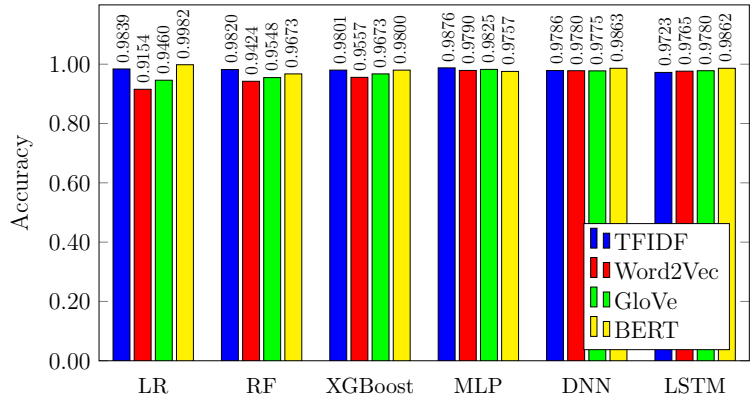

Recall that our embeddings approach consists of applying four different word embedding techniques, and
then applying the same six learning models as considered above to the resulting features
Figure~\ref{fig:bar_embed} shows the accuracy obtained with our embeddings approach.
 

From these results embeddings results, we observe that all of the embedding techniques are 
able to achieve very high accuracy for at least one model. Also, BERT and TFIDF both achieve consistently 
high accuracy for every model tested, whereas Word2Vec and GloVe lag somewhat when 
classic ML models are used.

\subsection{Ablation Study}

In this section, we consider the effects of modifications to the features. 
We consider two sets of experiments, one
based on the Random Forest classifier and one based on a linear Support Vector Machine (SVM).
In both cases, these models are trained and tested using the extracted 
features discussed in Section~\ref{sect:FA}
with the original dataset.

\subsubsection{Random Forest Model}

Table~\ref{tab:mod_features} shows the results of modifying individual features 
in the ChatGPT samples by
modifying the specified feature value by~$\hbox{}+10\%$ and by~$\hbox{}-10\%$,  
while keeping all other features unchanged.
Note that in each case, we test the resulting data using 
the Random Forest model that was trained on the unmodified data. 

We observe that decreasing the \texttt{lowercase\_letter\_ratio} 
has by far the most effect, making the prediction no better
than random. All other modifications have a minimal effect on the accuracy.


\begin{table}[!htb]
\centering
\caption{Feature modifications and Random Forest (original accuracy~0.9247)}\label{tab:mod_features}
\adjustbox{scale=0.85}{
\begin{tabular}{c|cc}
\toprule
\multirow{2}{*}{\textbf{Feature}} & \multicolumn{2}{c}{\textbf{Accuracy}} \\ \cline{2-3} \\[-2.5ex]
                                                   & \textbf{Increase~10\%} & \textbf{Decrease~10\%} \\ \midrule
 \texttt{lowercase\_letter\_ratio} &  0.9623 & 0.4955 \\ 
 \texttt{verb\_ratio} & 0.9182 & 0.9292 \\ 
 \texttt{average\_sentence\_length} & 0.9227 & 0.9194 \\ 
 \texttt{noun\_ratio} & 0.9129 & 0.9313 \\ 
 \texttt{negation\_ratio} &  0.9242 & 0.9250 \\ 
 \texttt{subjectivity} & 0.9281 & 0.9201 \\ 
 \texttt{sentence\_complexity} & 0.9258 & 0.9223 \\ 
 \texttt{homonym\_frequency} & 0.9265 & 0.9218 \\ 
 \texttt{burstiness} & 0.9237 & 0.9194 \\ \bottomrule
\end{tabular}
}
\end{table}

\subsubsection{Linear SVM Model}

As mentioned in Section~\ref{sect:SVM},
SVM training consists of constructing a hyperplane to separate the classes. 
We choose a linear SVM
since each feature has an associated well-defined weight. Training this model, 
we obtain an accuracy of~0.9234, which is comparable to the result for the Random Forest 
model discussed in the previous section.

As a first experiment with our SVM model, 
we consider the effect when each individual feature is altered by~$\hbox{}+10\%$ and~$\hbox{}-10\%$
then tested on our trained SVM model. These results are summarized in Table~\ref{tab:mod_features_SVM}.

\begin{table}[!htb]
\centering
\caption{Feature modification and SVM (original accuracy~0.9234)}\label{tab:mod_features_SVM}
\adjustbox{scale=0.85}{
\begin{tabular}{c|cc}
\toprule
\multirow{2}{*}{\textbf{Feature}} & \multicolumn{2}{c}{\textbf{Accuracy}} \\ \cline{2-3} \\[-2.5ex]
                                                   & \textbf{Increase~10\%} & \textbf{Decrease~10\%} \\ \midrule
 \texttt{verb\_ratio} & 0.9156 & 0.9295 \\ 
 \texttt{capital\_letter\_ratio} &  0.9266 & 0.9198 \\ 
 \texttt{lowercase\_letter\_ratio} &  0.9567 & 0.4620 \\ 
 \texttt{lexical\_diversity} &  0.8901 & 0.9404 \\ 
 \texttt{homonym\_frequency} & 0.9389 & 0.8903 \\ 
 \texttt{synonym\_frequency} & 0.9000 & 0.9363 \\ 
 \texttt{burstiness} & 0.9184 & 0.9278 \\
 \texttt{sentence\_count} & 0.9174 & 0.9282 \\ 
 \texttt{negation\_ratio} & 0.9229 & 0.9239 \\ 
 \texttt{word\_count} & 0.9303 & 0.9139 \\ \bottomrule
\end{tabular}
}
\end{table}

As with the Random Forest model, decreasing the 
\texttt{lowercase\_letter\_ratio} by~10\%\ has the effect of making the SVM model 
prediction essentially random, while other modifications have relatively little effect.
If an attacker is able to make appropriate modifications to the data, both the Random Forest
and the SVM model will be rendered useless. The next logical step would be to
train models on such modified data to determine how well we can distinguish
between human and modified-ChatGPT data.

In Table~\ref{tab:mod_features_SVM_retrain}, we give the results when the specified modification
is made to the ChatGPT data, and the SVM model is retrained on the modified dataset.
The most interesting case in Table~\ref{tab:mod_features_SVM_retrain} is when
the \texttt{lowercase\_letter\_ratio} is decreased by~10\%.

\begin{table}[!htb]
\centering
\caption{Feature modification and retrained SVM (original accuracy~0.9234)}
	\label{tab:mod_features_SVM_retrain}
\adjustbox{scale=0.85}{
\begin{tabular}{c|cc}
\toprule
\multirow{2}{*}{\textbf{Feature}} & \multicolumn{2}{c}{\textbf{Accuracy}} \\ \cline{2-3} \\[-2.5ex]
                                                   & \textbf{Increase~10\%} & \textbf{Decrease~10\%} \\ \midrule
 \texttt{verb\_ratio} & 0.9157 & 0.9297 \\ 
 \texttt{capital\_letter\_ratio} &  0.9265 & 0.9198 \\ 
 \texttt{lowercase\_letter\_ratio} &  0.9453 & 0.8125 \\ 
 \texttt{lexical\_diversity} &  0.8988 & 0.9393 \\ 
 \texttt{homonym\_frequency} & 0.9389 & 0.8962 \\ 
 \texttt{synonym\_frequency} & 0.9079 & 0.9343 \\ 
 \texttt{burstiness} & 0.9184 & 0.9279 \\
 \texttt{sentence\_count} & 0.9176 & 0.9282 \\ 
 \texttt{negation\_ratio} & 0.9229 & 0.9239 \\ 
 \texttt{word\_count} & 0.9289 & 0.9166 \\ \bottomrule
\end{tabular}
}
\end{table}

By retraining our SVM, 
we are able achieve an accuracy of~0.8125. Recall that for this same case, without
retraining the SVM, the prediction was essentially random, as can be 
seen in Table~\ref{tab:mod_features_SVM}. This result indicates that even
if the \texttt{lowercase\_letter\_ratio} is modified in this way, there is sufficient
statistical information available to distinguish between the classes with 
reasonable accuracy.

\begin{figure}[!htb]
    \centering
    \begin{tikzpicture}[scale=0.8]
\begin{axis}[ 
		   width=0.7\textwidth,
		   height=0.575\textwidth,
	 	   x tick label style={scale=0.85,
   		 	/pgf/number format/.cd,
			/pgf/number format/1000 sep={},
   			fixed,
   			fixed zerofill,
    			precision=0
		   },
		   x label style={scale=0.85},
	 	   y tick label style={scale=0.85,
    		 	/pgf/number format/.cd,
   			fixed,
   			fixed zerofill,
    			precision=2
		    },
		   y label style={scale=0.85},
                    xmin=-10,xmax=10,
                    ymin=0.48,ymax=1.0,
                    xtick={-10,-8,-6,-4,-2,0,2,4,6,8,10},
                    ytick={0.5,0.6,0.7,0.8,0.9,1.0},
                    xlabel={Percentage change},
                    ylabel={Accuracy},
                    legend pos=south east,
                    legend style={nodes={scale=0.85, transform shape},
                    	at={(1.55,0.45)},
                         anchor=south east,
                         }
                    ] 
\addplot[color=red,very thick,mark=none] coordinates { 
(-10, 0.9208392526871633)
(-9, 0.9212815289355885)
(-8, 0.921633339587745)
(-7, 0.9220220065939368)
(-6, 0.9222833516498244)
(-5, 0.922682070388935)
(-4, 0.9229032085131477)
(-3, 0.9231075937491624)
(-2, 0.9232148122336291)
(-1, 0.9233890422708875)
(0, 0.9234292492025625)
(1, 0.9234191974696437)
(2, 0.9234191974696437)
(3, 0.9234426515131209)
(4, 0.9234560538236792)
(5, 0.9234057951590854)
(6, 0.9232851743640603)
(7, 0.9231209960597206)
(8, 0.9229568177553811)
(9, 0.9227993406063205)
(10, 0.9227021738547726)
};
\addlegendentry{\texttt{verb\_ratio}}
\addplot[color=blue,very thick,mark=none] coordinates { 
(-10, 0.9234962607553542)
(-9, 0.9235364676870292)
(-8, 0.9235364676870292)
(-7, 0.9235532205752272)
(-6, 0.923506312488273)
(-5, 0.9234895596000751)
(-4, 0.9234795078671563)
(-3, 0.9234460020907604)
(-2, 0.9234057951590854)
(-1, 0.923425898624923)
(0, 0.9234292492025625)
(1, 0.9234024445814458)
(2, 0.9234057951590854)
(3, 0.923392392848527)
(4, 0.9233856916932479)
(5, 0.9232952260969791)
(6, 0.9232282145441875)
(7, 0.9231276972149999)
(8, 0.9230070764199748)
(9, 0.9229266625566248)
(10, 0.9227993406063205)
};
\addlegendentry{\texttt{capital\_letter\_ratio}}
\addplot[color=green,very thick,mark=none] coordinates { 
(-10, 0.5047712225587692)
(-9, 0.5143806792290991)
(-8, 0.5354256573833329)
(-7, 0.5726170691827271)
(-6, 0.6296439006084649)
(-5, 0.7008771812260434)
(-4, 0.7762551263837886)
(-3, 0.8425664084488166)
(-2, 0.8901211568874474)
(-1, 0.917046398799153)
(0, 0.9234292492025625)
(1, 0.912184710644115)
(2, 0.8860669579435495)
(3, 0.8499309781006246)
(4, 0.8075796767362693)
(5, 0.7626249765459565)
(6, 0.7182231216661753)
(7, 0.6780396440346316)
(8, 0.6423258369742944)
(9, 0.612395126919881)
(10, 0.5880933872999705)	
};
\addlegendentry{\texttt{lowercase\_letter\_ratio}}
\addplot[color=yellow,very thick,mark=none] coordinates { 
(-10, 0.905507009408422)
(-9, 0.9084655694641757)
(-8, 0.9114040260540918)
(-7, 0.9137896373334763)
(-6, 0.9159407081780899)
(-5, 0.9179276007183639)
(-4, 0.9195526308735626)
(-3, 0.9208627067306404)
(-2, 0.9218310236684805)
(-1, 0.9229099096684268)
(0, 0.9234292492025625)
(1, 0.9235900769292626)
(2, 0.9234795078671563)
(3, 0.9231612029913957)
(4, 0.9223470126249765)
(5, 0.9216366901653845)
(6, 0.9205108960784839)
(7, 0.9188356072586914)
(8, 0.9169425308923258)
(9, 0.9146272817433726)
(10, 0.911765888439167)
};
\addlegendentry{\texttt{lexical\_diversity}}
\addplot[color=black,very thick,mark=none] coordinates { 
(-10, 0.9107171576379768)
(-9, 0.913387568016726)
(-8, 0.9159708633768462)
(-7, 0.9179410030289222)
(-6, 0.9198608840164044)
(-5, 0.9213719945318573)
(-4, 0.9223403114696974)
(-3, 0.9229702200659393)
(-2, 0.9232684214758624)
(-1, 0.923586726351623)
(0, 0.9234292492025625)
(1, 0.9230037258423353)
(2, 0.9220856675690889)
(3, 0.9208057469107674)
(4, 0.9194956710536897)
(5, 0.9180281180475514)
(6, 0.9162657142091296)
(7, 0.9143223791781703)
(8, 0.9122450210416275)
(9, 0.9096449727933096)
(10, 0.9071018843648645)
};
\addlegendentry{\texttt{homonym\_frequency}}
\addplot[color=red,dashed,very thick,mark=none] coordinates { 
(-10, 0.912362291259013)
(-9, 0.9139739191036534)
(-8, 0.9155620929048168)
(-7, 0.9172976921221219)
(-6, 0.9184201356313828)
(-5, 0.9196330447369127)
(-4, 0.9205779076312757)
(-3, 0.9216031843889887)
(-2, 0.92256144959391)
(-1, 0.9231645535690353)
(0, 0.9234292492025625)
(1, 0.9234124963143646)
(2, 0.9233555364944916)
(3, 0.9231779558795936)
(4, 0.922916610823706)
(5, 0.9224341276436058)
(6, 0.9216735465194199)
(7, 0.9207119307368591)
(8, 0.9194052054574209)
(9, 0.9180549226686681)
(10, 0.9165404615755757)
};
\addlegendentry{\texttt{synonym\_frequency}}
\addplot[color=blue,dashed,very thick,mark=none] coordinates { 
(-10, 0.9221426273889619)
(-9, 0.9224274264883265)
(-8, 0.9225648001715496)
(-7, 0.9227725359852038)
(-6, 0.9228864556249498)
(-5, 0.9230405821963706)
(-4, 0.9232449674323854)
(-3, 0.9233756399603291)
(-2, 0.923392392848527)
(-1, 0.9233990940038063)
(0, 0.9234292492025625)
(1, 0.9234393009354813)
(2, 0.9234426515131209)
(3, 0.9234393009354813)
(4, 0.9232516685876645)
(5, 0.9233655882274104)
(6, 0.9233756399603291)
(7, 0.9233153295628166)
(8, 0.9232952260969791)
(9, 0.9232784732087812)
(10, 0.9232985766746187)
};
\addlegendentry{\texttt{burstiness}}
\addplot[color=green,dashed,very thick,mark=none] coordinates { 
(-10, 0.9213451899107407)
(-9, 0.9216132361219074)
(-8, 0.921931540997668)
(-7, 0.9222196906746724)
(-6, 0.922544696705712)
(-5, 0.9227624842522851)
(-4, 0.9230204787305332)
(-3, 0.9231478006808373)
(-2, 0.9232851743640603)
(-1, 0.923392392848527)
(0, 0.9234292492025625)
(1, 0.9234862090224355)
(2, 0.9234527032460396)
(3, 0.9235163642211918)
(4, 0.923626933283298)
(5, 0.9236001286621813)
(6, 0.9234929101777146)
(7, 0.923506312488273)
(8, 0.9233588870721312)
(9, 0.9232985766746187)
(10, 0.9233957434261667)	
};
\addlegendentry{\texttt{sentence\_count}}
\addplot[color=yellow,dashed,very thick,mark=none] coordinates { 
(-10, 0.9234862090224355)
(-9, 0.9234929101777146)
(-8, 0.923506312488273)
(-7, 0.9235230653764709)
(-6, 0.9235096630659125)
(-5, 0.9234661055565979)
(-4, 0.9234527032460396)
(-3, 0.9234594044013188)
(-2, 0.9234828584447958)
(-1, 0.9234728067118771)
(0, 0.9234292492025625)
(1, 0.9233990940038063)
(2, 0.9233957434261667)
(3, 0.9233957434261667)
(4, 0.9233856916932479)
(5, 0.9233488353392125)
(6, 0.9233722893826896)
(7, 0.9233521859168521)
(8, 0.9233722893826896)
(9, 0.9233789905379688)
(10, 0.9233655882274104)	
};
\addlegendentry{\texttt{negation\_ratio}}
\addplot[color=black,dashed,very thick,mark=none] coordinates { 
(-10, 0.9225748519044683)
(-9, 0.922779237140483)
(-8, 0.9230673868174873)
(-7, 0.9231947087677915)
(-6, 0.923271772053502)
(-5, 0.9233555364944916)
(-4, 0.9234359503578417)
(-3, 0.9235297665317501)
(-2, 0.9234460020907604)
(-1, 0.9234527032460396)
(0, 0.9234292492025625)
(1, 0.9233890422708875)
(2, 0.9232349156994666)
(3, 0.9229534671777414)
(4, 0.9226452140348996)
(5, 0.9222699493392661)
(6, 0.9218209719355617)
(7, 0.9213552416436593)
(8, 0.9208761090411987)
(9, 0.9203768729729005)
(10, 0.9200083094325462)
};
\addlegendentry{\texttt{word\_count}}
\end{axis}
\end{tikzpicture}
    \caption{SVM individual feature modification\hspace*{1.05in}}\label{fig:linearSVM}
\end{figure}
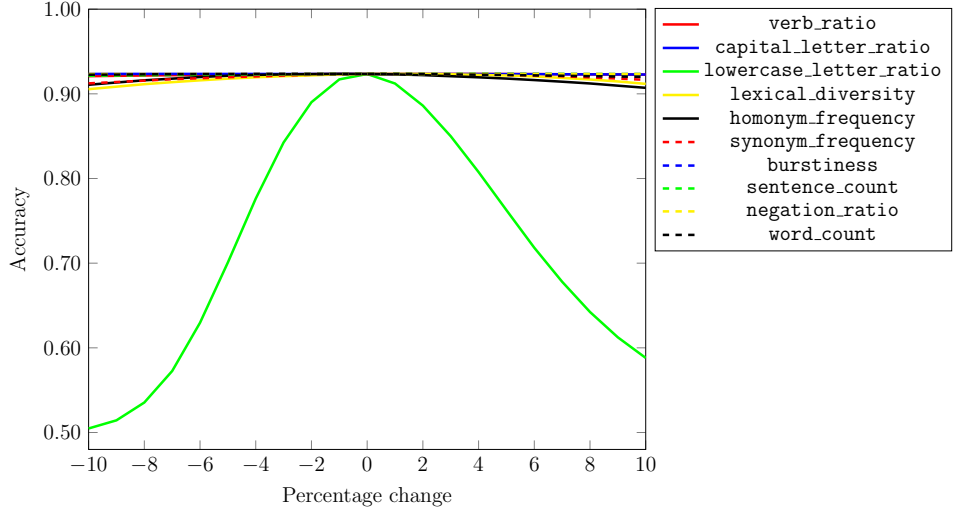

As a final experiment, we directly modify the weights of our trained linear
SVM to determine the robustness of the model itself.
In Figure~\ref{fig:linearSVM}, we give the results when 
each individual feature weight is modified from~$\hbox{}-10\%$ to~$\hbox{}+10\%$.

Analogous to the data modifications discussed above, we observe that
only the feature weight associated with the \texttt{lowercase\_letter\_ratio}
has a substantial effect, at least within the range of~$\hbox{}-10\%$ to~$\hbox{}+10\%$.
This shows that the linear SVM model is robust with respect to changes to
the weights, and further emphasizes the overriding importance of the
\texttt{lowercase\_letter\_ratio} to the success of the model.


%
%

\section{Conclusion and Future Work}\label{conclusion}

The rapid advancement of generative AI in general, and Large Language Models like GPT-3 in particular,
presents both opportunities and challenges. One of the challenges is in distinguishing 
between human-generated and machine-generated text. In the research presented in this paper, 
we considered two methodologies to address this challenge---one based on feature analysis and another
based on advanced word embedding techniques.

Through feature analysis, we identified sentence patterns and general tendencies of the GPT model. 
First, we computed a variety of statistical features of text. By employing  
classic machine learning techniques (Linear Regression, Random Forrest, XGBoost) 
and deep learning architectures (MLP, DNN, LSTM), we identified the how these features 
can be used to differentiate between human and GPT-generated texts. In addition to these techniques, 
we also explored various dimensionality reduction methods (PCA, LDA, RF, Lasso). 
We found that based on the features that we extracted from text, we could distinguish
chatbot from human with an accuracy of better than~0.96. 

Since the chatbot-generated text tends to be slightly longer than the human-generated text,
we also explored the effect of normalizing the length on our feature analysis approach. 
Surprisingly, this resulted in improved accuracy, with the best models achieving perfect separation
on our test set.

We then applied word embedding techniques (TFIDF, Word2Vec, GloVe, BERT), 
which are designed to capture semantic aspects of the language.
The same machine learning and deep learning models were trained on the
resulting word embedding sequences, and we obtained accuracies that were
better than for our original feature analysis experiments. In the best cases, we obtained
accuracies in excess of~0.99.

Finally, we consider an ablation study to determine the effect of modification to features
on our feature analysis based models. We found that relatively small modifications to 
the \texttt{lowercase\_letter\_ratio} feature have a 
profound impact on the accuracies of models---to the point where the models
only marginally outperform a coin flip---while modifications to other features had
minimal effect. This result indicates that there is considerable scope 
for improvement, with respect to making chatbot-generated text more human-like.

For future work, it would be interesting to expand the scope of experimentation to include a 
broader range of model architectures, features, word embeddings, and dimensionality reduction techniques,
along with more extensive tuning of model hyperparameters. This could involve experimenting with stacked models, 
which use a hierarchical approach to refine predictions through successive layers of processing,
as well as various ensemble techniques.

Experiments involving other chatbots should be considered. With recent advances, new 
models are constantly being created, including newer versions of the GPT model considered in this paper. 
It would be interesting to rank the ``humanness'' of the text generated by various chatbots.

Finally, it would be very interesting to post-process chatbot-generated text using the insights gained from 
our research, with the goal of making the modified text more human-like. We have provided evidence
that such modified text can defeat detection techniques that are based on feature analysis of the original chatbot data,
but it is unclear how effective such techniques will be for models trained on word embeddings.

\bibliographystyle{plain}
\bibliography{references.bib}

\appendix
\makeatletter
\renewcommand\section{\@startsection {section}{1}{\z@}%
                                     {-3.5ex \@plus -1ex \@minus -.2ex}%
                                     {2.3ex \@plus.2ex}%
                                     {\noindent\normalfont\Large\bfseries Appendix }}
\makeatother

\section{}\label{appendix_a}

Here, we provide a few examples of human text and the corresponding ChatGPT-generated text.

\subsection*{Example 1}

\noindent{\textbf{Human:}
Sell Yourself First. Before you embark on selling fine art online, it is crucial to establish a strong personal brand and present yourself as a credible artist. Begin by creating a captivating artist statement that reflects your artistic vision, inspirations, and unique style. Craft a compelling biography that highlights your artistic journey, education, exhibitions, and any accolades or awards received. Showcase your artwork through a well-designed portfolio website, social media platforms, and online galleries. Engage with your audience by sharing your creative process, behind-the-scenes glimpses, and meaningful stories behind your artworks. By selling yourself first and establishing a strong online presence, you will attract art enthusiasts who appreciate your talent and are more likely to purchase your fine art.}

\bigskip

\noindent{\textbf{ChatGPT:}
Before doing anything else, stop and sum up yourself as an artist. Now, think about how to translate that to an online profile. Be it the few words Twitter allows you or an entire page of indulgence that your own website would allow you. Bring out the most salient features of your creativity, your experience, your passion, and your reasons for painting. Make it clear to readers why you are an artist who loves art, produces high quality art, and is a true champion of art. If you're not great with words, find a friend who can help you with this really important aspect of selling online---the establishment of your credibility and reliability.}

\subsection*{Example 2}

\noindent{\textbf{Human:}
Reading the classics before 1600 is a profound way to enhance your literary knowledge and truly become well-read. Dive into the works of ancient Greek and Roman writers, such as Homer's epic poems ``The Iliad'' and ``The Odyssey,'' which provide invaluable insights into heroic quests and the human condition. Explore the philosophical musings of Plato and Aristotle, whose ideas shaped Western thought and continue to resonate today. Engage with the plays of Sophocles, Euripides, and Aeschylus, witnessing the timeless tragedies that delve into the depths of human emotions and moral dilemmas. Additionally, immerse yourself in medieval literature like Dante Alighieri's ``The Divine Comedy,'' a compelling journey through Hell, Purgatory, and Heaven, offering profound theological and philosophical reflections. By delving into these classics before 1600, you will open doors to a wealth of wisdom, artistry, and cultural understanding.}

\bigskip

\noindent{\textbf{ChatGPT:}
Reading the classics is the very first thing you have to do to be well-read. If you want to build a solid foundation for your understanding of the books you read, then you can't avoid some of the earliest plays, poems, and oral tales ever written down. Remember that the novel didn't really get popular until the 18th century, so you won't find novels on this list. Without reading the poetry of Homer or the plays of Sophocles, you won't be able to call yourself well-read. Here's a list to get you started: The Epic of Gilgamesh (Unknown author) (18th--17th century BCE) The Iliad and The Odyssey by Homer (850--750 BCE, 8th century BCE) ``The Oresteia'' by Aeschylus (458 BCE) Oedipus the King by Sophocles (430 BCE) Medea by Euripides (431 BCE) Aeneid by Virgil (29--19 BCE) One Thousand and One Nights (Unknown author) (700–1500) Beowulf (Unknown author) (975--1025) The Tale of Genji by Murasaki Shikibu (11th century) The Divine Comedy by Dante (1265--1321) The Decameron by Boccaccio (1349–53) The Canterbury Tales by Chaucer (14th century).}

\subsection*{Example 3}

\noindent{\textbf{Human:}
While you certainly don't have to be wealthy to adopt a child, you will have to be able to demonstrate that you have the space to house the child and the financial means to support the child. Home and community environment are also important. Talk to your family and friends, and evaluate the multicultural resources available in your area. Make sure those close to you accept your decision to adopt a child of another race. You also want to make sure there are no legal restrictions that would prevent you from adopting a child. If you're thinking about adopting internationally, you want to look at the laws in the child's home country as well.}

\bigskip

\noindent{\textbf{ChatGPT:}
Before embarking on the journey to adopt a child of another race, it is crucial to honestly assess your ability to care for a child. Consider factors such as your emotional readiness, financial stability, and available support systems. Reflect on your understanding of the unique challenges that may arise from transracial adoption, including potential racial identity issues and the importance of cultural preservation. Assessing your ability to provide a loving and inclusive environment will help you choose an agency that specializes in transracial adoption and can offer the necessary resources and support to ensure a successful and fulfilling adoption experience.}

\subsection*{Example 4}

\noindent{\textbf{Human:}
Most computers are set to boot into the hard drive first, which means you will need to change some settings to boot from your newly-burned CD or USB. Start by rebooting the computer. Once the computer reboots, press the key used to enter the boot menu. The key for your system will be displayed on the same screen as the manufacturer's logo. Typical keys include~F12, F2, or~Del. For Windows 8 users, hold the Shift key and click restart. This will load the Advanced Startup Options, where you can boot from CD. If your computer doesn't give you direct access to the boot menu from the manufacturer's splash screen, it's most likely hidden in the BIOS menu. You can access the BIOS menu in the same way that you would get to the boot menu. At the manufacturer splash screen, the key should be listed in one of the bottom corners. Once you're in the boot menu, select your live CD or USB. Once you've changed the settings, save and exit the BIOS setup or boot menu. Your computer will continue with the boot process.}

\bigskip

\noindent{\textbf{ChatGPT:}
To boot into the Live CD or Live USB for the Installing Any Linux Distribution section, begin by inserting the Linux installation media into your computer's CD/DVD drive or USB port. Restart your computer and access the BIOS settings by pressing the designated key during startup (typically F2, F12, or Del). Once in the BIOS, navigate to the Boot menu and prioritize the boot order to allow the computer to boot from the installation media. Save the changes and exit the BIOS. As the computer restarts, it will load the Linux distribution from the Live CD or Live USB, presenting you with a graphical interface where you can explore the distribution, test its features, and ultimately proceed with the installation process.}

\subsection*{Example 5}

\noindent{\textbf{Human:}
Start the first lift. Do the bench press- first do three sets of five and then three sets of three. As you progress through sets, try to go up in weight. This is how you get stronger. Second lift: Do the dumbbell bench press- first, three sets of eight and then three sets of five. Third lift: Use the chest press machine- three sets of eight or three sets of five. This machine is typically more tiring. Fourth lift: Do wide grip push ups- five sets of ten. Give yourself very little rest time for this lift. Fifth lift: Use the lat pull down machine---do sets of five until failure. Sixth lift: Do pull-ups/chin-ups---do three sets of max. Seventh lift: Kneeling dumbbell row---five sets of eight. Eighth lift: Use the rowing machine---three sets of eight or three sets of five. Ninth lift: Bicep curls drop sets---three sets of seven reps per arm. Alternate arms. Tenth lift: Hammer curls drop sets---three sets of seven reps per arm. Alternate arms. Eleventh lift: Curling bar curls---three sets of eight.}

\bigskip

\noindent{\textbf{ChatGPT:}
In order to get strong for football during the off season, it's important to establish a well-rounded strength training routine. One effective approach is to dedicate Tuesdays and Fridays as chest, bicep, and back days. Start by warming up with some light cardio exercises to get the blood flowing. Then, focus on chest exercises such as bench presses, push-ups, and dumbbell flies to target your pectoral muscles. Next, move on to bicep exercises like dumbbell curls, hammer curls, and chin-ups to strengthen the muscles in your arms. Finally, wrap up the workout with back exercises such as lat pulldowns, rows, and deadlifts to enhance your upper body strength and posture. Remember to maintain proper form and gradually increase the weight or resistance as you progress. By incorporating these specific workouts on Tuesdays and Fridays, you can effectively build strength and power for football in the off season.}


\section{}\label{appendix_b}

In this appendix, we provide feature analysis graphs for each of the~30 features.

\begin{figure}[!htb]
    \centering
    \includegraphics[scale=0.80]{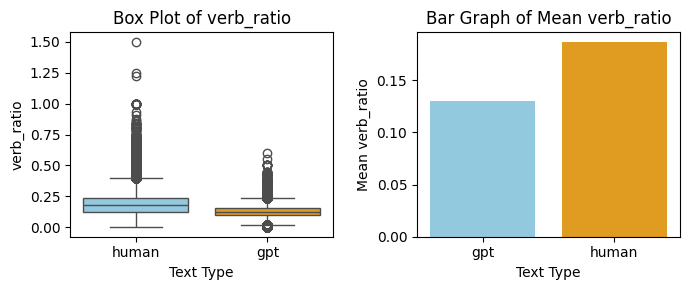}
    \caption{Verb ratio}\label{fig:verb_ratio}
\end{figure}

\begin{figure}[!htb]
    \centering
    \includegraphics[scale=0.80]{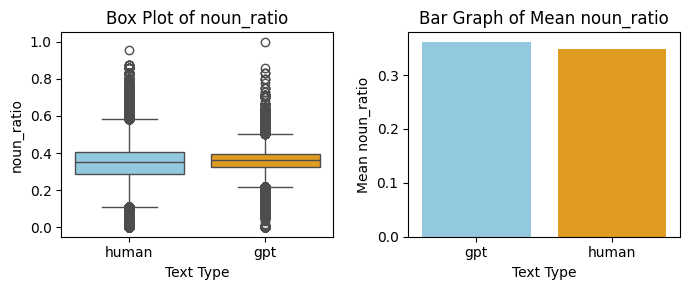}
    \caption{Noun ratio}\label{fig:noun_ratio}
\end{figure}

\begin{figure}[!htb]
    \centering
    \includegraphics[scale=0.80]{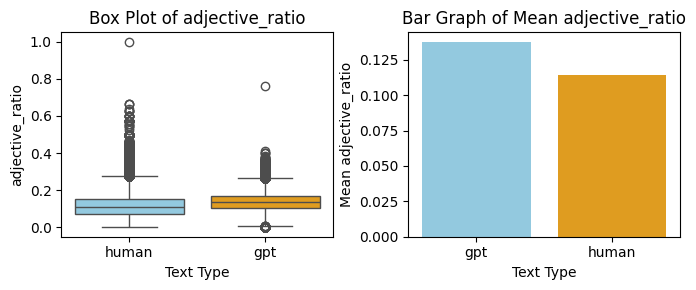}
    \caption{Adjective ratio}\label{fig:adjective_ratio}
\end{figure}

\begin{figure}[!htb]
    \centering
    \includegraphics[scale=0.80]{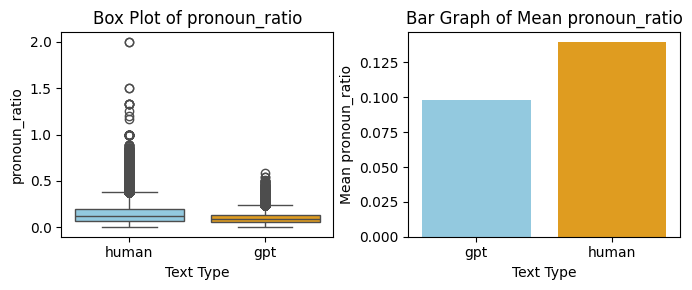}
    \caption{Pronoun ratio}\label{fig:pronoun_ratio}
\end{figure}

\begin{figure}[!htb]
    \centering
    \includegraphics[scale=0.80]{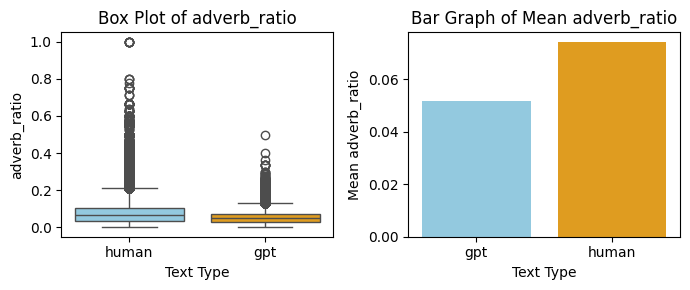}
    \caption{Adverb ratio}\label{fig:adverb_ratio}
\end{figure}

\begin{figure}[!htb]
    \centering
    \includegraphics[scale=0.80]{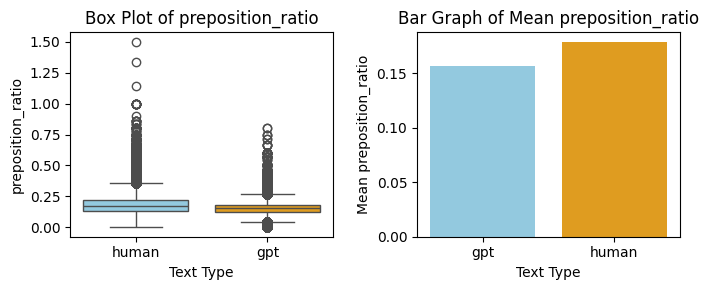}
    \caption{Preposition ratio}\label{fig:preposition_ratio}
\end{figure}

\begin{figure}[!htb]
    \centering
    \includegraphics[scale=0.80]{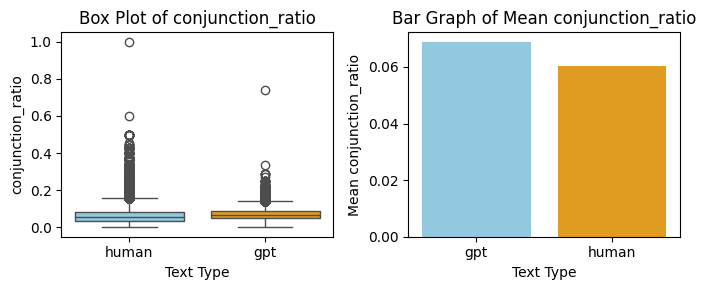}
    \caption{Conjunction ratio}\label{fig:conjunction_ratio}
\end{figure}

\begin{figure}[!htb]
    \centering
    \includegraphics[scale=0.80]{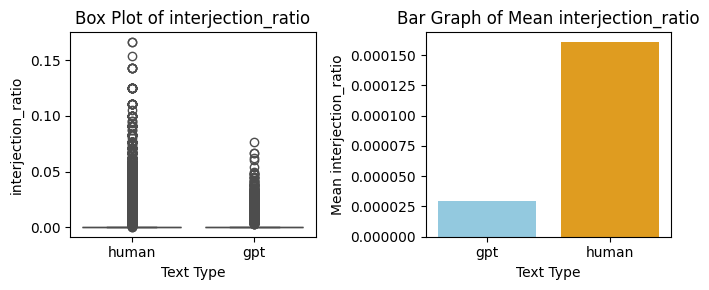}
    \caption{Interjection ratio}\label{fig:interjection_ratio}
\end{figure}

\begin{figure}[!htb]
    \centering
    \includegraphics[scale=0.80]{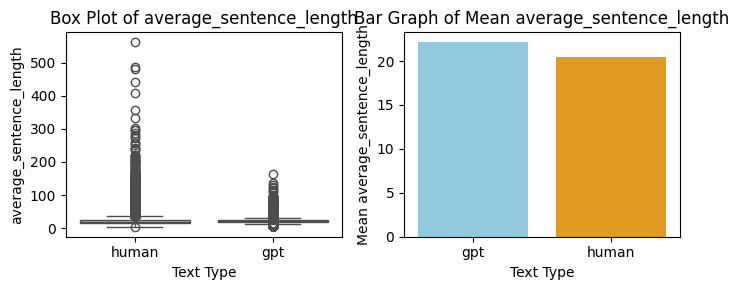}
    \caption{Average sentence length}\label{fig:average_sentence_length}
\end{figure}

\begin{figure}[!htb]
    \centering
    \includegraphics[scale=0.80]{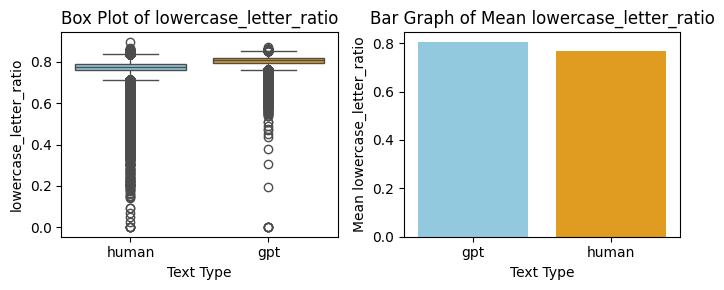}
    \caption{Lowercase letter ratio}\label{fig:lowercase_letter_ratio}
\end{figure}

\begin{figure}[!htb]
    \centering
    \includegraphics[scale=0.80]{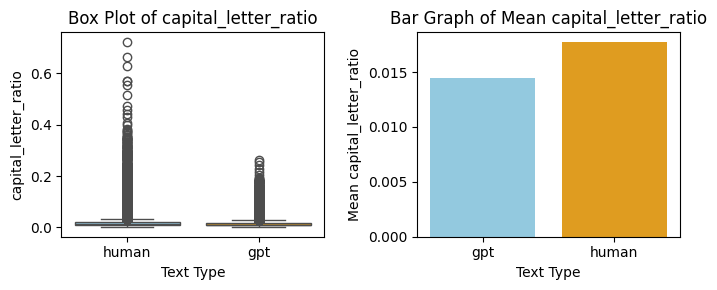}
    \caption{Capital letter ratio}\label{fig:capital_letter_ratio}
\end{figure}

\begin{figure}[!htb]
    \centering
    \includegraphics[scale=0.80]{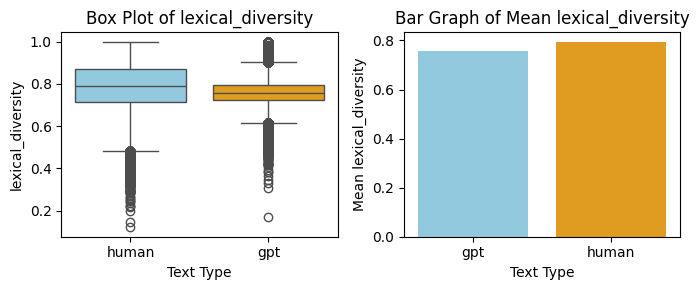}
    \caption{Lexical diversity}\label{fig:lexical_diversity}
\end{figure}

\begin{figure}[!htb]
    \centering
    \includegraphics[scale=0.80]{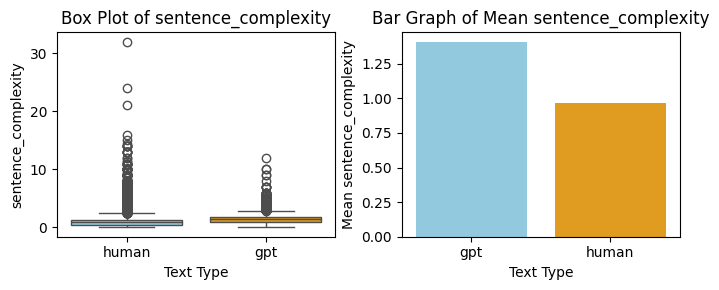}
    \caption{Sentence complexity}\label{fig:sentence_complexity}
\end{figure}

\begin{figure}[!htb]
    \centering
    \includegraphics[scale=0.80]{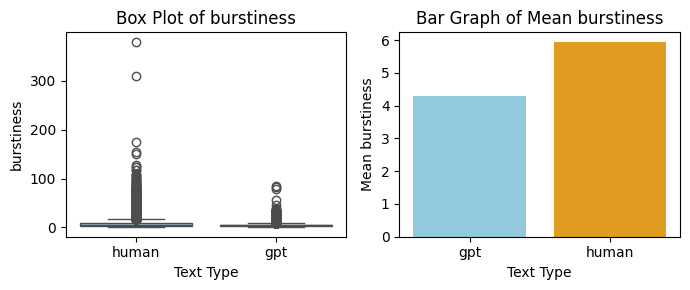}
    \caption{Burstiness}\label{fig:burstiness}
\end{figure}

\begin{figure}[!htb]
    \centering
    \includegraphics[scale=0.80]{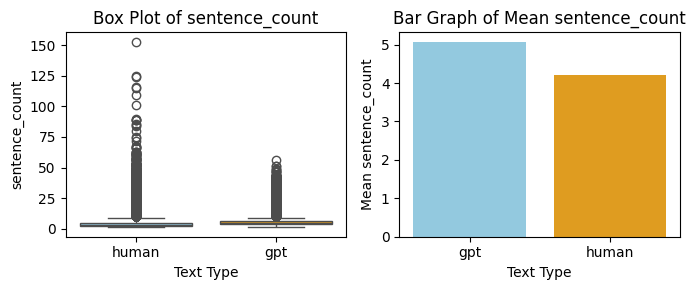}
    \caption{Sentence count}\label{fig:sentence_count}
\end{figure}

\begin{figure}[!htb]
    \centering
    \includegraphics[scale=0.80]{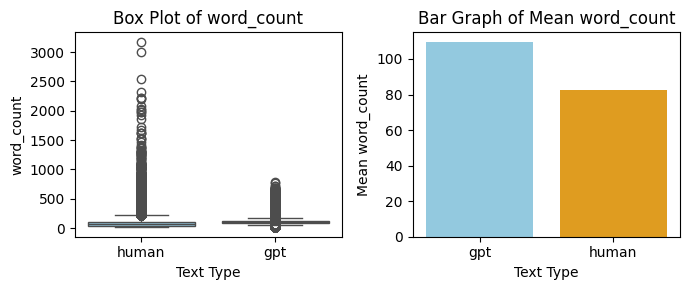}
    \caption{Word count}\label{fig:word_count}
\end{figure}

\begin{figure}[!htb]
    \centering
    \includegraphics[scale=0.80]{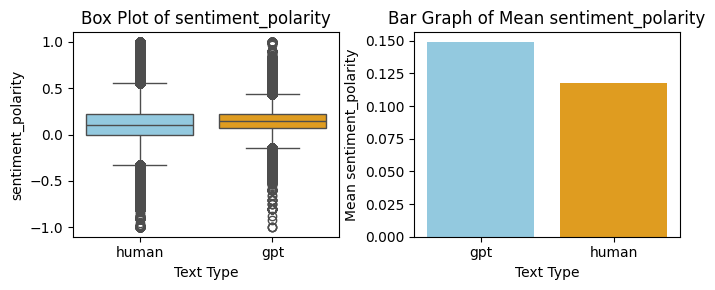}
    \caption{Sentiment polarity}\label{fig:sentiment_polarity}
\end{figure}

\begin{figure}[!htb]
    \centering
    \includegraphics[scale=0.80]{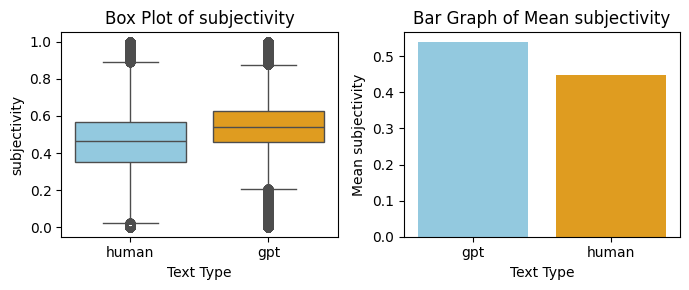}
    \caption{Subjectivity}\label{fig:subjectivity}
\end{figure}

\begin{figure}[!htb]
    \centering
    \includegraphics[scale=0.80]{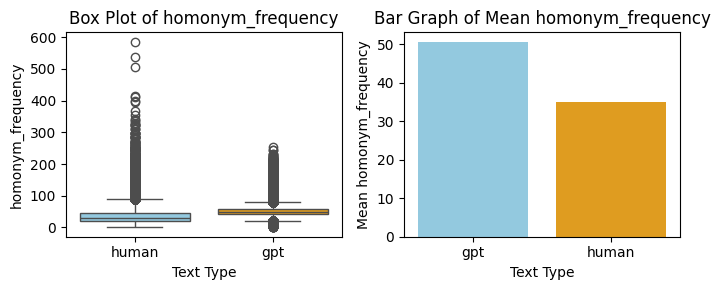}
    \caption{Homonym frequency}\label{fig:homonym_frequency}
\end{figure}

\begin{figure}[!htb]
    \centering
    \includegraphics[scale=0.80]{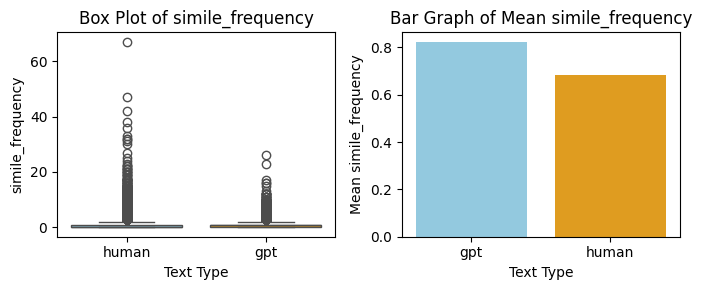}
    \caption{Simili frequency}\label{fig:simili_frequency}
\end{figure}

\begin{figure}[!htb]
    \centering
    \includegraphics[scale=0.80]{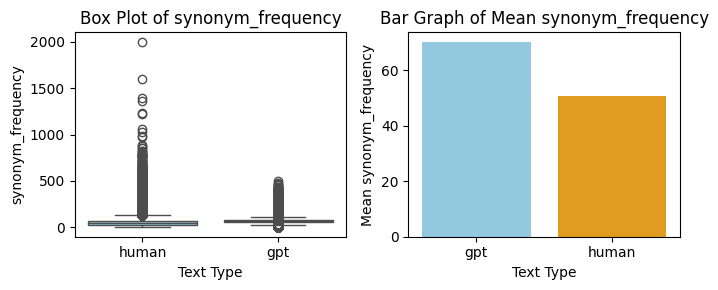}
    \caption{Synonym frequency}\label{fig:synonym_frequency}
\end{figure}

\begin{figure}[!htb]
    \centering
    \includegraphics[scale=0.80]{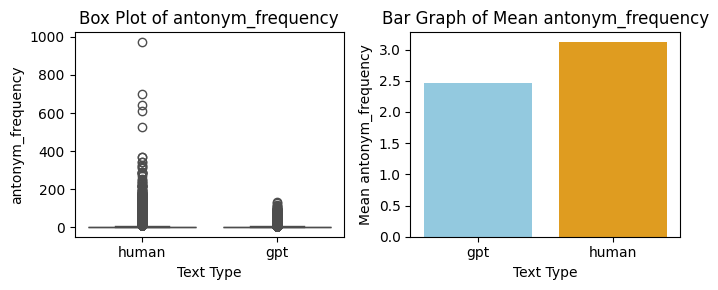}
    \caption{Antonym frequency}\label{fig:antonym_frequency}
\end{figure}

\begin{figure}[!htb]
    \centering
    \includegraphics[scale=0.80]{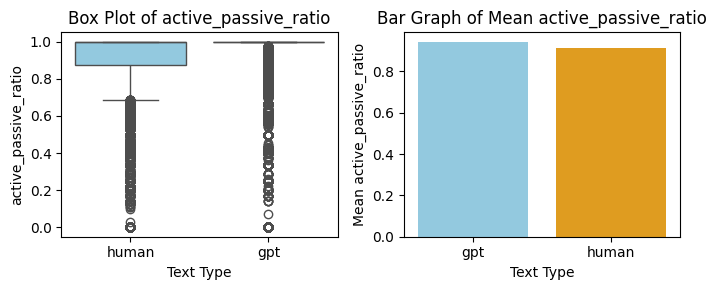}
    \caption{Active passive ratio}\label{fig:active_passive_ratio}
\end{figure}

\begin{figure}[!htb]
    \centering
    \includegraphics[scale=0.80]{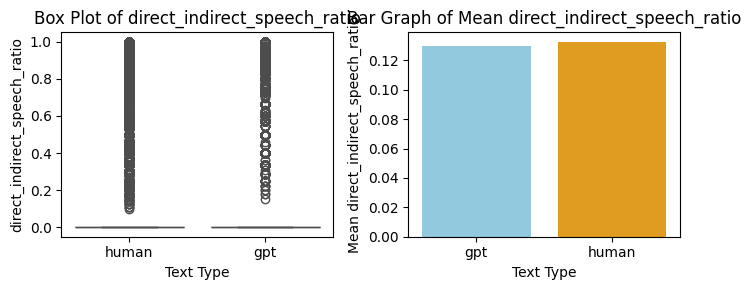}
    \caption{Direct indirect speech ratio}\label{fig:direct_indirect_speech_ratio}
\end{figure}

\begin{figure}[!htb]
    \centering
    \includegraphics[scale=0.80]{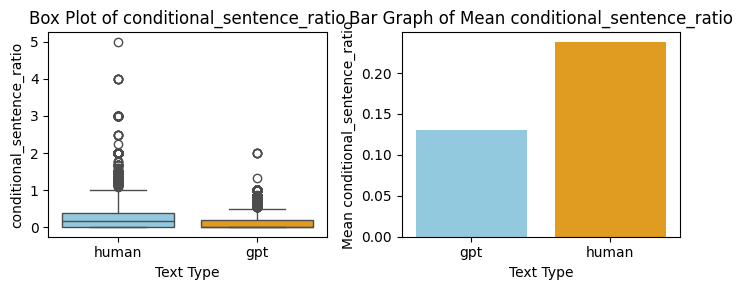}
    \caption{Conditional sentence ratio}\label{fig:conditional_sentence_ratio}
\end{figure}

\begin{figure}[!htb]
    \centering
    \includegraphics[scale=0.80]{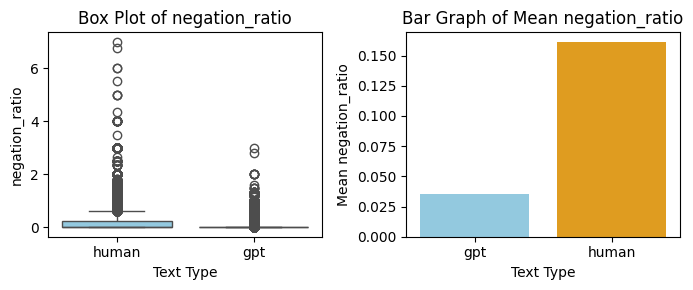}
    \caption{Negation ratio}\label{fig:negation_ratio}
\end{figure}

\begin{figure}[!htb]
    \centering
    \includegraphics[scale=0.80]{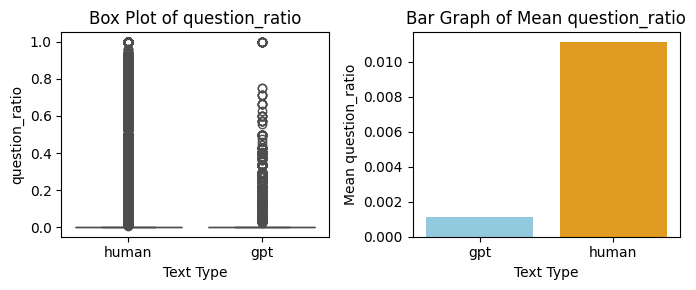}
    \caption{Question ratio}\label{fig:question_ratio}
\end{figure}

\begin{figure}[!htb]
    \centering
    \includegraphics[scale=0.80]{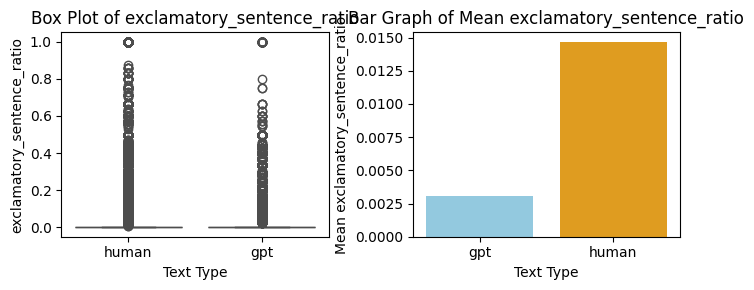}
    \caption{Exclamatory sentence ratio}\label{fig:exclamatory_sentence_ratio}
\end{figure}

\begin{figure}[!htb]
    \centering
    \includegraphics[scale=0.80]{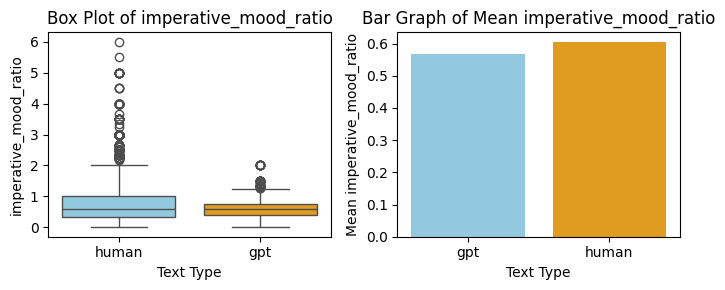}
    \caption{Imperative mood ratio}\label{fig:imperative_mood_ratio}
\end{figure}

\begin{figure}[!htb]
    \centering
    \includegraphics[scale=0.80]{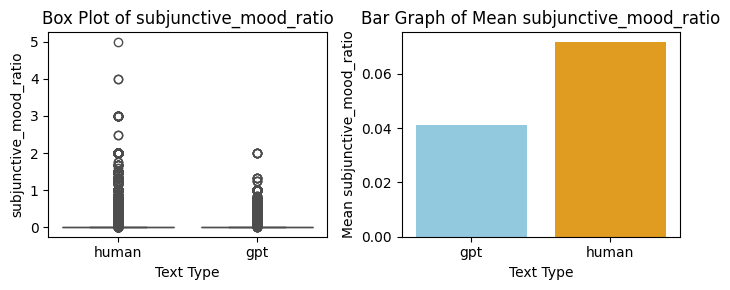}
    \caption{Subjunctive mood ratio}\label{fig:subjunctive_mood_ratio}
\end{figure}

\end{document}